\documentclass[10pt,twocolumn,letterpaper]{article}

\usepackage{iccv}
\usepackage{times}
\usepackage{epsfig}
\usepackage{graphicx}
\usepackage{amsmath}
\usepackage{amssymb}

\usepackage[breaklinks=true,bookmarks=false]{hyperref}

\iccvfinalcopy 

\def\iccvPaperID{****} 

\begin{document}

\title{Pruning for feature preserving circuits in CNNs}

\author{
Chris Hamblin \thanks{email for correspondence. Code available at \url{https://github.com/chrishamblin7/circuit_explorer}} 
\qquad\qquad\qquad   Talia Konkle
\qquad\qquad\qquad   George Alvarez \\
\texttt{chrishamblin@fas.harvard.edu  tkonkle@fas.harvard.edu   alvarez@wjh.harvard.edu} \\
Harvard University, Department of Psychology \\
}


\maketitle
\ificcvfinal\thispagestyle{empty}\fi

\begin{abstract}

Modern deep learning techniques have dramatically changed computer vision approaches, replacing hand-engineered image processing algorithms with large neural networks trained end-to-end. However, where hand-engineered algorithms are built by composing simpler, human understandable subfunctions, the inner-workings of deep networks are notoriously opaque. In this work we present \textit{circuit pruning}, a saliency-based approach for decomposing neural networks into sparse subnetworks, each of which computes the activations of a targeted latent feature. We demonstrate how our technique can be used to identify a subnetwork that preserves feature activations in general, across many image inputs, or alternatively isolate yet sparser \textit{'subcircuits'} responsible for a feature's activation to particular images. Concretely, we show how subcircuits combine to generate polysemantic features, and construct a shape detector from simpler parts. Broadly, we offer that pruning for feature-preserving circuits constitute a powerful approach to mechanistic interpretability, revealing the implicitly learned modular structure routing through the hierarchical layers of deep neural networks. \par  

\end{abstract}

\section{Introduction}\label{sec:intro}

Understanding the latent features learned in deep neural networks has been the subject of much research. Many techniques have been developed for characterizing \textit{what} latent features represent, such as viewing the output activation maps of a cnn filter \cite{Yosinski_deepviz_toolbox}, the attention maps in a visual transformer\cite{dino}, or viewing dataset examples or feature visualizations \cite{deepviz,olah2017feature} that maximally excite/inhibit a latent feature.  Other work shows that latent features can correspond to semantic annotations in densely segmented image sets \cite{datasetade,datasetcontext,datasetintrinsic,datasetparts,datasettextures}, even when those annotations are not part of the objective function on which the network was trained \cite{netdissect}. In the case of self-supervised learning, the emergence of useful latent features is often the explicit goal of training the network itself \cite{simCLR,barlow-twins,dino,self-supervised-transfer}. \par 
However, these approaches lack mechanistic interpretabilty, providing no insight into how these sophisticated features are constructed, or how latent features are linked across layers, or whether different activations across different images arise through similar or distinct computations.  Mechanistic interpretability is difficult, as in typical architectures the function that computes any particular feature from pixels is entangled with all the others. In the general case, a feature corresponds to a direction, \(\mathbf{\hat{f}}\), in a given layer's vector space. Thus, the function that computes it requires first transforming an image into that space, then taking the dot product with the feature direction; 
\begin{equation}\label{eq: feature function}
f(\mathbf{x}) = \mathcal{F}_{l}(\mathbf{x})\cdot \mathbf{\hat{f}}. 
\end{equation}
Where \(\mathcal{F}_{l}\) corresponds to the first \(l\) layers of the neural network \textit{in their entirety}. With \(\mathcal{F}_{l}\) in its functional description, how are we to possibly understand the function \(f(\mathbf{x})\)?\par 

\begin{figure}[t]
  \centering

   \includegraphics[width=0.8\linewidth]{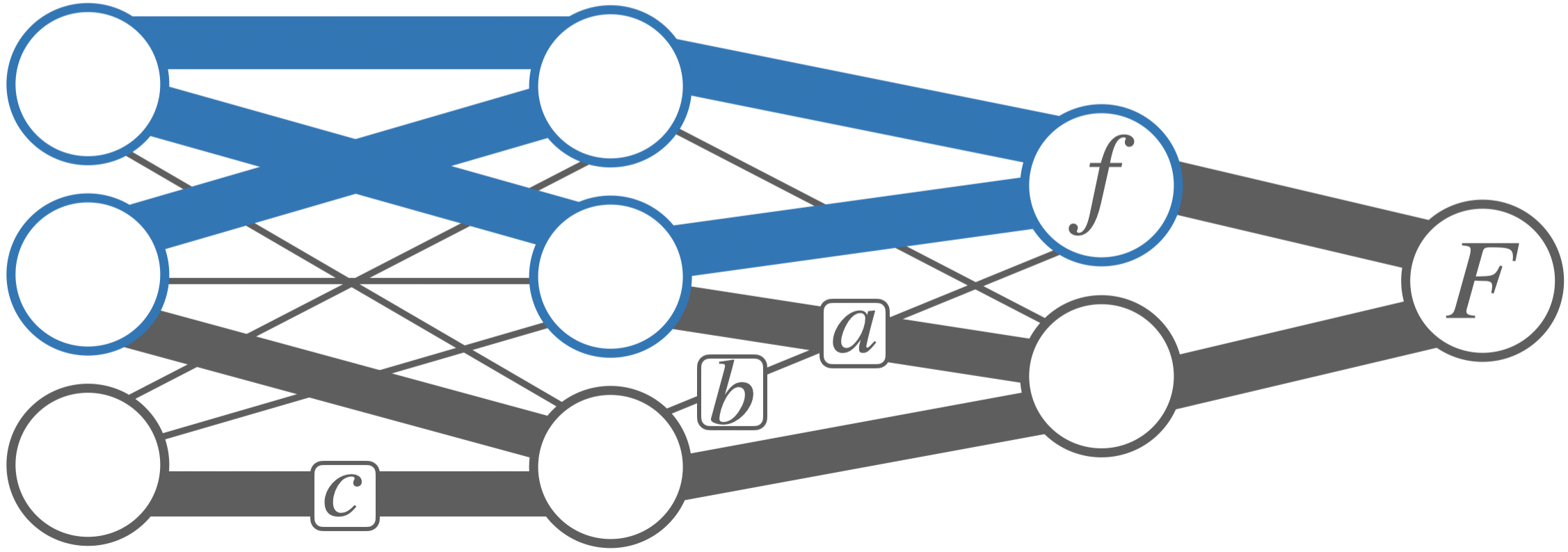}

   \caption{A network for computing the objective \(F\), where the edge width corresponds to weight magnitude. The blue edges/nodes represent a sparse, multi-layer, feature-preserving circuit for \(f\). Connection \(a\) can be excluded as it is not connected to \(f\), and connection \(b\) can be excluded due to its low magnitude. Connection \(c\) cannot be excluded by such an obvious heuristic, but a saliency score based on gradients from \(f\) could identify \(c\) for exclusion, as gradients from \(f\) to \(c\) must pass through \(b\).}
   \label{fig:demo}
\end{figure}

Here, we hypothesize that after training there exists a sparse path of computations through the network, \(\hat{f}(\mathbf{x})\), such that \(\hat{f}(\mathbf{x}) \approx f(\mathbf{x})\). We solve for \(\hat{f}(\mathbf{x})\) by leveraging techniques from saliency-based neural network pruning. Further, we offer that analyzing these feature-preserving circuits provides insight into the modular structures implicitly learned in a neural network, which constitute sparse routes through the full computation graph. 

\pagebreak

We highlight our main contributions as follows:

\begin{itemize}
  \item We demonstrate how saliency-based pruning can be used to modularize a network into feature-wise circuits. We test the efficacy of several saliency criteria towards this end, and characterize the distribution of sparsities over features that naturally emerges in Imagenet trained models.
  \item By pruning structurally with respect to convolutional kernels, our technique can be used to automate the generation of mechanistically interpretable circuit diagrams like those in Cammarata et al. (2020) \cite{circuitsthread}, which outline the hierarchical image filtering process that transform images into feature activations. 
  \item Our technique can identify image-wise \textit{subcircuits}, decomposing the circuit that computes a feature into yet sparser modules, each responsible for the feature's activation to different inputs. Concretely, we identify subcircuits that decompose polysemantic features, and decompose a circle detector into exhitatory and inhibitory parts. 

\end{itemize}

\section{Related Work}\label{sec:related work}

This work sits at the intersection of three lines of research, that on mechanistic interpretability, modular neural networks, and neural network pruning. Mechanistic interpretability refers to the process of reverse-engineering human-understandable algorithms from a network, endeavoring to understand not merely \textit{what} the network represents, but \textit{how} it forms such representations by operating on inputs \cite{gilpin2019explaining,interpretability_survey}. Cammarata et. al. (2020) \cite{circuitsthread,circuit_curve,circuit_weightviz,circuit_equivariance} provide a promising template for how to mechanistically interpret convolutional neural networks. By viewing feature visualizations in conjunction with the kernels that connect them, the authors provide strong visual intuition for how simple features are combined into complex ones. The authors describe many interesting 'circuits' they’ve discovered in InceptionV1 \cite{inception}, from early circuits encoding curves and basic shapes, to late-layer circuits encoding real-world objects. Many of the design choices underlying our method were made with these circuits in mind (hence the name 'circuit pruning'); our method can be viewed as \textit{automating} and \textit{validating} the identification of such circuits. 

For example, a prototypical circuit from their work shows how a car feature is constructed with three kernels spatially arranging features in the previous layer for windows, wheels, and car bodies \cite{circuitszoomin}. Of course, when one 'zooms-in' on these particular kernels,  they are implicitly ignoring all the other connections that lead into the putative 'car' feature. What justifies the choice of connections, and how much has the original 'car' feature's activation profile been preserved in this simplified circuit? Furthermore, even if we accept the 'car' feature is built by way of the 'wheel', 'window', and 'car body' features alone, how are these constituent features built? We've pushed our question of mechanistic car detection back by one layer; it would be useful to \textit{extend} circuit diagrams backwards layer-wise towards the pixel inputs. Of course such a diagram runs the risk of being massive and unparsable without considerable exclusions. Our circuit pruning technique automates the identification of these hierarchical circuits, while simultaneously quantifying feature preservation at different levels of sparsity. \par

Our technique can be viewed as a form of network modularization, i.e. a technique for dividing a network into simpler subfunctions. Many approaches to modularization have been proposed in the literature; for example, one can build modules explicitly into a network by way of a general purpose branching architecture \cite{branch_multicolumn,branch_wang2015multi,branching_pc_dropout}. Alternatively one could designate modules for subtasks of the objective that are known a priori {\cite{neural_module_explain,neural_module_networks}, but this is limited to certain problem domains \cite{visual_question_answering, referential_expression_grounding}. Related methods train a designated \textit{controller} in parallel with a conventional network that learns to route inputs through it \cite{conditional_computation,moe_routing,module_learning,routingnetworks}.  Other work on modularization is post hoc, specifying modules in a standard network architecture with dense connections after it has been trained. This has been done in a manner agnostic to the function of the modules by weighted graph clustering \cite{spectral_clustering_1,spectral_clustering_2,watanabe2019interpreting}. Other techniques have specified per-class modules in image classification networks \cite{modules_inout_1,modules_inout_2,modules_by_weight_mask, data_routing_paths_1, data_routing_paths_2}, or modules per input dimension \cite{watanabe2019interpreting}. The technique we present here is for post-hoc, feature-wise modularization. \par

\par 
\section{Methods}\label{sec:methods}

\subsection{Pruning Problem Statement}\label{sec:problem}

We endeavor to simplify the function that computes a given feature, \(f(\mathbf{x})\), by removing parameters from the network, while minimally affecting the feature's activations. We can formulate our \textit{circuit pruning} problem similarly to conventional pruning; given a latent feature in a network parameterized by \(\boldsymbol{\theta} = \{\theta\}_{i=1}^m\) and a set of input images \(\mathcal{D}=\left\{\mathbf{x}_{i}\right\}_{i=1}^{n}\), we want to identify a subnetwork parameterized by \(\boldsymbol{\bar{\theta}}\) under a sparsity constraint, such that the change in \(f\)'s responses to \(\mathcal{D}\) is minimized:

\begin{equation} \label{eq:problem form 1}
\begin{gathered}
\underset{\boldsymbol{\bar{\theta}}}{\arg \min } \: \Delta f(\boldsymbol{\bar{\theta}},\boldsymbol{\theta} ; \mathcal{D}):= \sum_{i=1}^{n}\left|f\left(\boldsymbol{\theta} ;\mathbf{x}_{i}\right) -f\left(\boldsymbol{\bar{\theta}} ;\mathbf{x}_{i}\right)\right| \\
\text { s.t. } \quad \left\|\boldsymbol{\bar{\theta}}\right\|_{0} \leq \kappa , \quad \bar{\theta}_{j} \in \{\theta_{j},0\} \,\, \forall j \in \{1 \ldots m\}
\end{gathered}
\end{equation}

 \(\boldsymbol{\bar{\theta}}\) corresponds to a sparse path of computations through the full network that generates an approximation of the target feature, with no more than \(\kappa\) parameters. The constraint that \( \bar{\theta}_{j} \in \{\theta_{j},0\} \,\, \forall j \in \{1 \ldots m\}\) ensures that we cannot fine-tune the network, that circuits we extract are latent within the network \textit{as is}. This additional constraint is particular to our goals, usually pruning a network involves iteratively removing then finetuning network parameters.
\par
As is typical with pruning problems, finding an optimal \(\boldsymbol{\bar{\theta}}\) cannot be brute forced, as there exist \(m\choose\kappa\) possible parameterizations. Fortunately, many saliency-based approaches to pruning have a similar problem statement to (eq \ref{eq:problem form 1}) in that they endeavor to find a sparse \(\boldsymbol{\bar{\theta}}\) that minimally changes the cost function \(\mathcal{L}\) of the original model. Such methods consider quickly-computable, parameter-wise \textit{saliency criteria} \(\boldsymbol{S}(\theta_{j})\) meant to approximate \(|\Delta\mathcal{L}|\) induced by removing individual parameters. With a simple substitution of \(f\) for \(\mathcal{L}\), these saliency criteria should be well suited for our problem.  We can then replace (eq \ref{eq:problem form 1}) with the simpler problem of identifying a parameterization \(\boldsymbol{\bar{\theta}}\) with maximal cumulative saliency;

\begin{equation} \label{eq:problem form saliency}
\begin{gathered}
\underset{\boldsymbol{\bar{\theta}}}{\arg \max } \: \boldsymbol{S}_{sum}(\boldsymbol{\bar{\theta}}) :=  \sum_{j\in supp(\boldsymbol{\bar{\theta}})}\boldsymbol{S}(\theta_{j}) \\
\text{ s.t. } \quad \left\|\boldsymbol{\bar{\theta}}\right\|_{0} \leq \kappa , \quad \bar{\theta}_{j} \in \{\theta_{j},0\} \,\, \forall j \in \{1 \ldots m\}
\end{gathered}
\end{equation}

Where solving problem (eq \ref{eq:problem form 1}) is hard, solving problem (eq \ref{eq:problem form saliency}) is easy, just choose parameters \(\theta_{j}\) with the top-\(\kappa\) \(\boldsymbol{S}(\theta_{j})\) for inclusion in \(\boldsymbol{\bar{\theta}}\).
\par

\subsection{Structured Pruning of Kernels}

It is sometimes practical to prune \textit{structurally}, i.e. with respect to architecturally related groups of parameters, rather than individual weights/biases. In such cases, saliency-based pruning logic follows the description above, but \(\theta_{j}\) refers to an individual structure, rather than an individual weight/bias. In the case of CNNs, the relevant structures are convolutional \textit{kernels} and \textit{filters}. As these terms are sometimes used interchangably in the literature, let us be clear in our terminology/notation. The \(l\)th convolutional layer takes a stack of \textit{activation maps} as input, \(\boldsymbol{A}^{l-1} \in \mathbb{R}^{C_{\text {in}} \times H_{\text {in }} \times W_{\text {in }}}\) and outputs the activation maps \(\mathbf{A}^{l} \in \mathbb{R}^{C_{\text {out}}\times H_{\text {out}} \times W_{\text {out}}}\). The layer is composed of \(C_{\text {out}}\) \textit{filters} (one for each output channel), each parameterized by weights \(\mathbf{w}^{l}_{c_{\text{out}}} \in \mathbb{R}^{C_{\text {in}} \times K_{h} \times K_{w}}\) and bias \(b^{l}_{c_{\text{out}}} \in \mathbb{R}\). Each filter transforms input activation maps to output maps by;

\begin{equation} \label{eq:filter eq}
\begin{gathered}
\mathbf{A}^{l}_{c_{\text{out}}} = \sum_{c_{\text{in}=1}}^{C_{\text {in}}} \mathbf{w}^{l}_{c_{\text{out}}, c_{\text{in}}} * \boldsymbol{A}^{l-1}_{c_{\text{in}}}+b^{l}_{c_{\text{out}}},
\end{gathered}
\end{equation}

where \(*\) denotes the convolution. A filter is composed of \(C_{\text{in}}\) \textit{kernels} with weights \(\mathbf{w}^{l}_{c_{\text{out}},c_{\text{in}}} \in \mathbb{R}^{K_{h} \times K_{w}}\) that compute the kernel-wise activation map \(\mathbf{A}^{l}_{c_{\text{out}},c_{\text{in}}} \in \mathbb{R}^{H_{\text {out}}\times W_{\text {out}}}\), before the filter sums these maps element-wise and adds its bias. Where individual activations in kernel-wise and filter-wise activation maps can be uniquely specified by \(a^{l}_{c_{\text{out}},c_{\text{in}},h,w}\) and \(a^{l}_{c_{\text{out}},h,w}\) respectively, we will often omit this subscripting for clarity when the context is unambiguous.

 Structured pruning of CNNs typically removes entire convolutional filters \cite{actgrad,thinet,filterentropy,filterentropy,ganpruning,filteredge,autofilter,softfilter,globalfilter,oraclefilter}, as this leads to models that are resource efficient when implemented with BLAS \cite{BLAS} libraries. Our goal is not resource efficiency, but rather to extract interpretable circuits from the model, and as such we will prune with respect to individual convolutional \textit{kernels}, which has received less treatment in the literature \cite{kernel_pruning_interpret,kernel_pruning,kernel_regularization,kernelexplore}. Kernels constitute the edges of our desired circuit diagrams (Figure \ref{fig:small_circuit_diagram}), thus an extracted circuit with kernel sparsity should make for a maximally parsable diagram.

\begin{figure*}
  \centering
  \includegraphics[width=\textwidth]{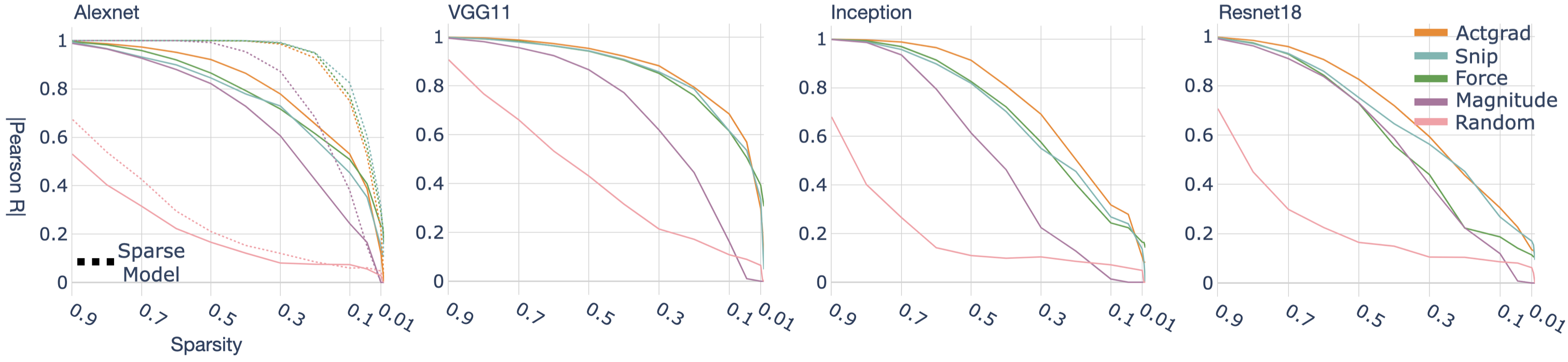}
  \caption{Median feature preservation scores, as a function of sparsities, plotted across models for different pruning methods.}
  \label{fig:comparison}
\end{figure*}

\subsection{Saliency Criteria}\label{sec:saliency}

We will test the efficacy of 3 saliency criteria from the literature for feature-preserving circuit pruning; ActGrad \cite{actgrad}, SNIP \cite{mozer_prune,snip}, and FORCE \cite{FORCE}. Each of these criteria, \(\mathbf{S}(\theta_{j})\), was originally formulated to approximate \(|\Delta\mathcal{L}|\) induced by removing \(\theta_{j}\), given the goal of identifying a subnetwork that sparsely achieves the objective function. A full justification of this approximation for each criterion is given in the appendix. By substituting \(f\) for \(\mathcal{L}\), we define analogous feature-wise saliency criteria below.\linebreak\linebreak
\textbf{ActGrad:}  This criterion was originally designed for the structured pruning of filters and computed with respect to a filter's activation map. Here, we can define an analogous kernel-wise criterion with respect to a kernel's activation map;
\begin{equation}\label{eq:actgrad kernel}
\boldsymbol{S}_{actgrad}(\theta_{j};f,\mathbf{x}) := \frac{1}{H_{\text{out}}W_{\text{out}}}\sum_{w=1}^{W_{\text{out}}}\sum_{h=1}^{H_{\text{out}}}|a_{h,w}\frac{\partial f(\mathbf{x})}{\partial a_{h,w}}|
\end{equation}
\textbf{SNIP:}  This criterion was originally designed for weight pruning. Here, we define a kernel-wise criterion by averaging across the weights in a kernel;
\begin{equation} \label{eq:kernel snip}
\boldsymbol{S}_{snip}(\theta_{j};f,\mathbf{x}) := \frac{1}{K_{w}K_{h}}\sum_{w=1}^{K_{w}}\sum_{h=1}^{K_{h}}|\text{w}_{h,w}\frac{\partial f(\mathbf{x})}{\partial \text{w}_{h,w}}|
\end{equation}
\textbf{FORCE:} This method implements a modification of SNIP, as the authors observed that \(\boldsymbol{S}_{snip}(\theta_{j})\) approximates \(\theta_{j}\)'s importance in the full network, not its importance in the desired sparse network. They posit that importance in a \(\kappa_{T}\) sparse network can be approximated by iteratively computing \(\boldsymbol{S}_{snip}(\boldsymbol{\theta})\) and pruning to \(\kappa_{i}\) sparsity for \(T\) iterations, such that \(\kappa_{i+1} < \kappa_{i} \: \forall i\) (without finetuning in between).\linebreak\linebreak
\textbf{Magnitude:} As a baseline method, we compare the above criteria to one based on the magnitude of kernel weights alone;

\begin{equation} \label{eq:magnitude}
\boldsymbol{S}_{mag}(\theta_{j};f) := \frac{1}{K_{w}K_{h}}\sum_{w=1}^{K_{w}}\sum_{h=1}^{K_{h}}|\text{w}_{h,w}|
\end{equation}
subject to the constraint that \(\boldsymbol{S}_{mag}(\theta_{j};f) = 0\) for kernels causally disconnected from \(f\) (such as kernels in a deeper layer than the target feature in a feed-forward network).\par
Besides the baseline magnitude method, each of the above criteria is image-conditional, and designed to identify important kernels for computing \(f(\mathbf{x}_{i})\) specifically. To identify important kernels for computing \(f\)'s response to multiple images \(\mathcal{D} = \{\mathbf{x}_{i}\}_{i=1}^{n}\), we simply sum the per-image scores;
\(\boldsymbol{S}(\boldsymbol{\theta};f,\mathcal{D}) = \sum_{i=1}^{n}\boldsymbol{S}(\boldsymbol{\theta};f,\mathbf{x}_{i})\).

\section{Circuit Pruning Efficacy}\label{sec:methodcompare}

All of the following experiments were conducted with a single NVIDIA \textit{GeForce RTX 2080 Ti} GPU. All feature visualizations in this work were generated using the \textit{Lucent} library \cite{lucent} (under an Apache License 2.0), a PyTorch implementation of \textit{Lucid} \cite{olah2017feature}, using the default hyper-parameters. 
\par
We tested the efficacy of the above saliency criteria by pruning circuits across several Imagenet \cite{imagenet_cvpr09} trained convolutional models, namely Alexnet \cite{alexnet}, VGG11 \cite{vgg}, Inception \cite{inception}, and Resnet18 \cite{resnet}. We trained an additional Alexnet model with hierarchical group sparsity regularization \cite{hierarch_reg}, which encourages groups of parameters (kernels and filters) towards zero magnitude (\textit{appendix}). Given our goal to identify the sparsest possible feature-preserving circuits, we reasoned it is useful to start with a model that already sparsely achieves the objective. \par
What features should we target in these pruning experiments? So far, we have characterized a feature in general terms, as our saliency criteria are well defined for any intermediate value computed by a network with accessible gradients. In these experiments we will constrain our notion of `feature' to mean the activations returned by a single convolutional filter, as such features are often analysed in interpretability research. \par

\begin{figure}[t]
  \centering

   \includegraphics[width=0.8\linewidth]{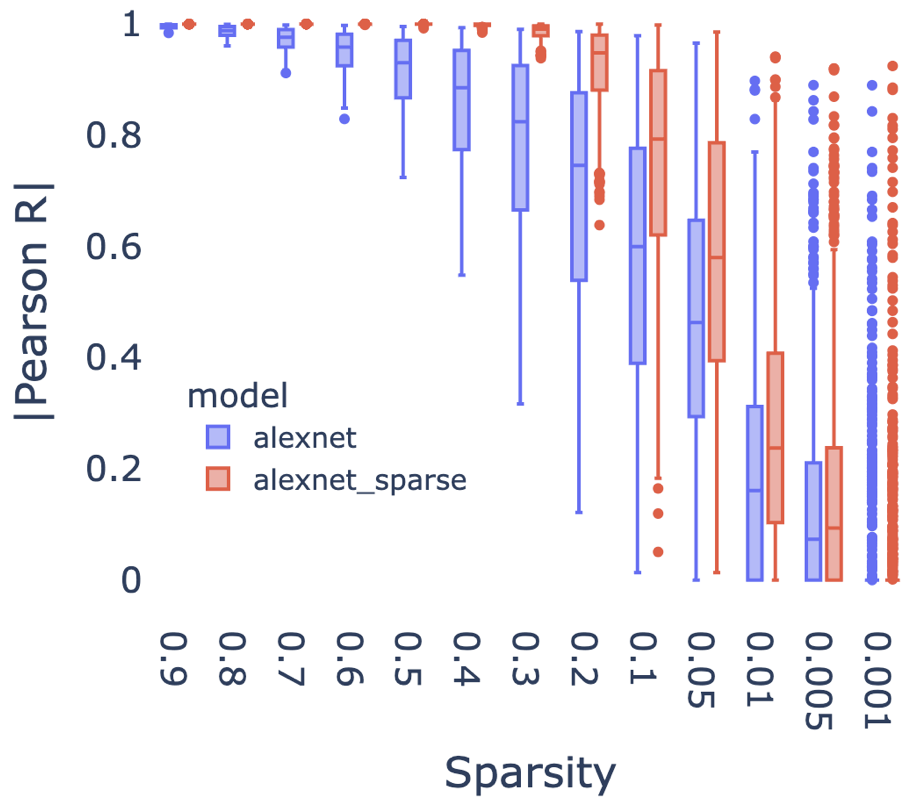}

   \caption{Distribution of feature preservation for 896 Alexnet features across sparsities.}
   \label{fig:full_distribution}
\end{figure}

\begin{figure*}[t]
  \centering
  \includegraphics[width=\textwidth]{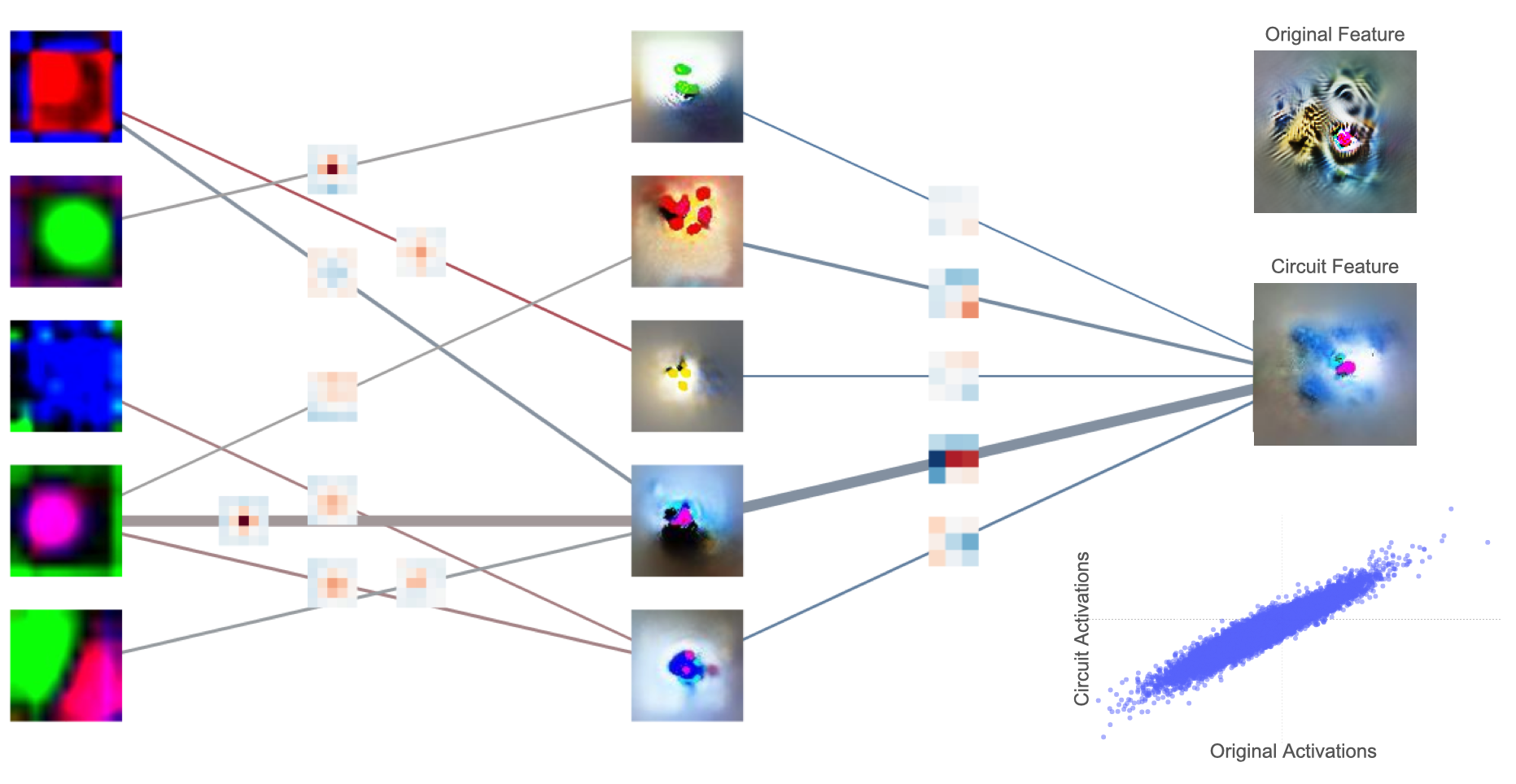}

  \caption{An \textit{extended circuit diagram} of feature \textit{conv3:56} in Alexnet. The bottom right panel shows the correlation between original activations and sub-circuit activations across 2000 images.}
  \label{fig:small_circuit_diagram}
\end{figure*}

For each model, we extracted circuits for 60 random convolutional filters, sampled evenly across the convolutional layers from the third convolutional layer
to the last. We excluded the first two convolutional layers from this sample, as we wanted to test whether our methods could identify sparse circuits with hierarchical structure. For each filter, we computed saliency scores with respect to 2000 random images from Imagenet (2 images per class). As convolutional filters return an activation map per image, \(f(\mathbf{x}) = \mathbf{A}_{f}\), rather than a scalar value, we use the gradients with respect to \(\sum|\mathbf{A}_{f}|)\) when computing each saliency criteria for this experiment. Using these saliency scores, we then extracted circuits at 13 sparsity levels\footnote{Throughout this paper, sparsity is reported with respect to \textit{relevant} kernels. That is, a .1 sparse circuit contains only 10\% of those kernels causally connected to the target feature.}, preserving 99\%-.1\% of the kernels with highest saliency. For this experiment, our measure of feature preservation was a correlation metric. Specifically, we computed the absolute value of the Pearson correlation between the original feature activations to the 2000 images and the corresponding pruned circuit activations. Circuits that score high on this metric have the same activation profile as the original feature, such that the original activations can be trivially recovered from the circuit activations. This measure was computed for circuits at each level of sparsity, across all 60 selected filters in each model. 
\par
Figure \ref{fig:comparison} shows how well feature activations were preserved (\(|\)Pearson's R\(|\)), as a function of circuit sparsity, with separate lines for the three pruning methods, as well as the magnitude baseline method and a further random pruning baseline. Across models and sparsities, \textit{ActGrad}, \textit{SNIP} and \textit{FORCE} pruning perform comparably, with slight improvements using the \textit{ActGrad} method in some regimes. We note that the three gradient-based methods outperform the baseline magnitude saliency criterion, even in the case of the sparsity regularized network. This empirically validates the intuitions behind figure \ref{fig:demo}; there are fewer kernels in a layer important for computing a given feature than those important for computing the objective function. These results provide support for the idea that the latent features learned in the later stages of deep neural networks are well approximated by sparse paths of computation through the network. 

While Figure \ref{fig:comparison} reflects the median feature preservation score across 60 filters, we observed that the circuits within these samples are far from homogeneously sparse. That is, some circuits maintain a high correlation with the full-model feature counterpart, even at extreme sparsities, while other features degrade quickly as more kernels are pruned. To get the full picture of the distribution of feature sparsities in a given model, our next analysis moved beyond the 60 features tested for methods comparison, and extracted circuits for every filter in the last 3 convolutional layers of Alexnet (896 features). Given its superior performance in the preceding experiment, we move forward with the \textit{actgrad} saliency citerion for all subsequent experiments.  Figure \ref{fig:full_distribution} shows the distributions of correlations for the original and sparsity regularized Alexnet. All features tested are preserved (Pearson's R \(> .99\)) at sparsities up to ~50\% in the regularized model. At higher sparsities (\(\geq 20\%\)) there is increasingly high variance in the correlations across features, but individual features can be identified that are well preserved at even the highest sparsity tested (.1\% of relevant kernels kept). 

\begin{figure*}
    \centering
    \begin{minipage}{0.49\textwidth}
        \centering 
        \includegraphics[width=\textwidth]{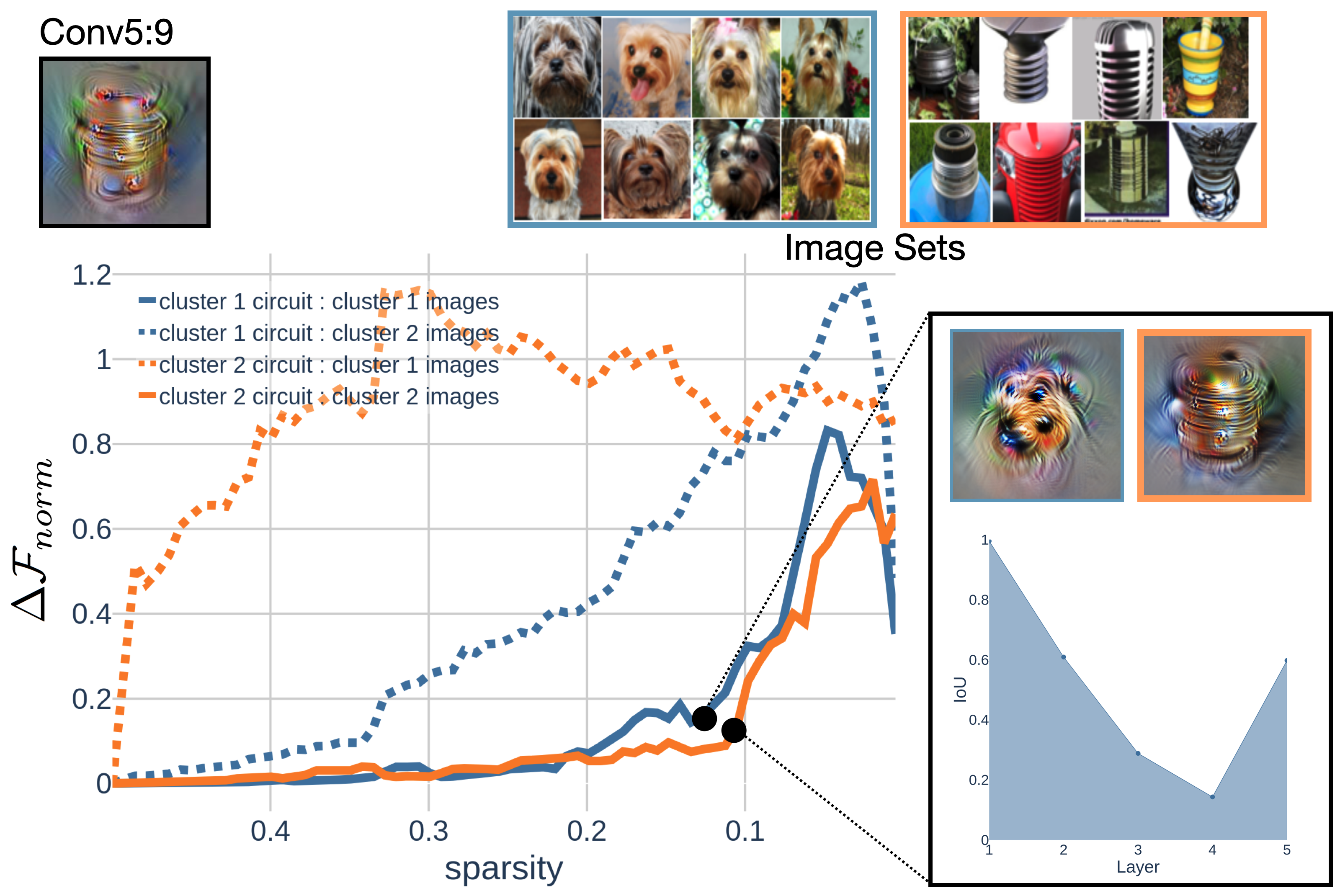} 
        \caption{terrier/metal subcircuits computed from clustered images.}\label{fig:poly clusters}
    \end{minipage}\hfill
    \begin{minipage}{0.49\textwidth}
        \centering
        \includegraphics[width=\textwidth]{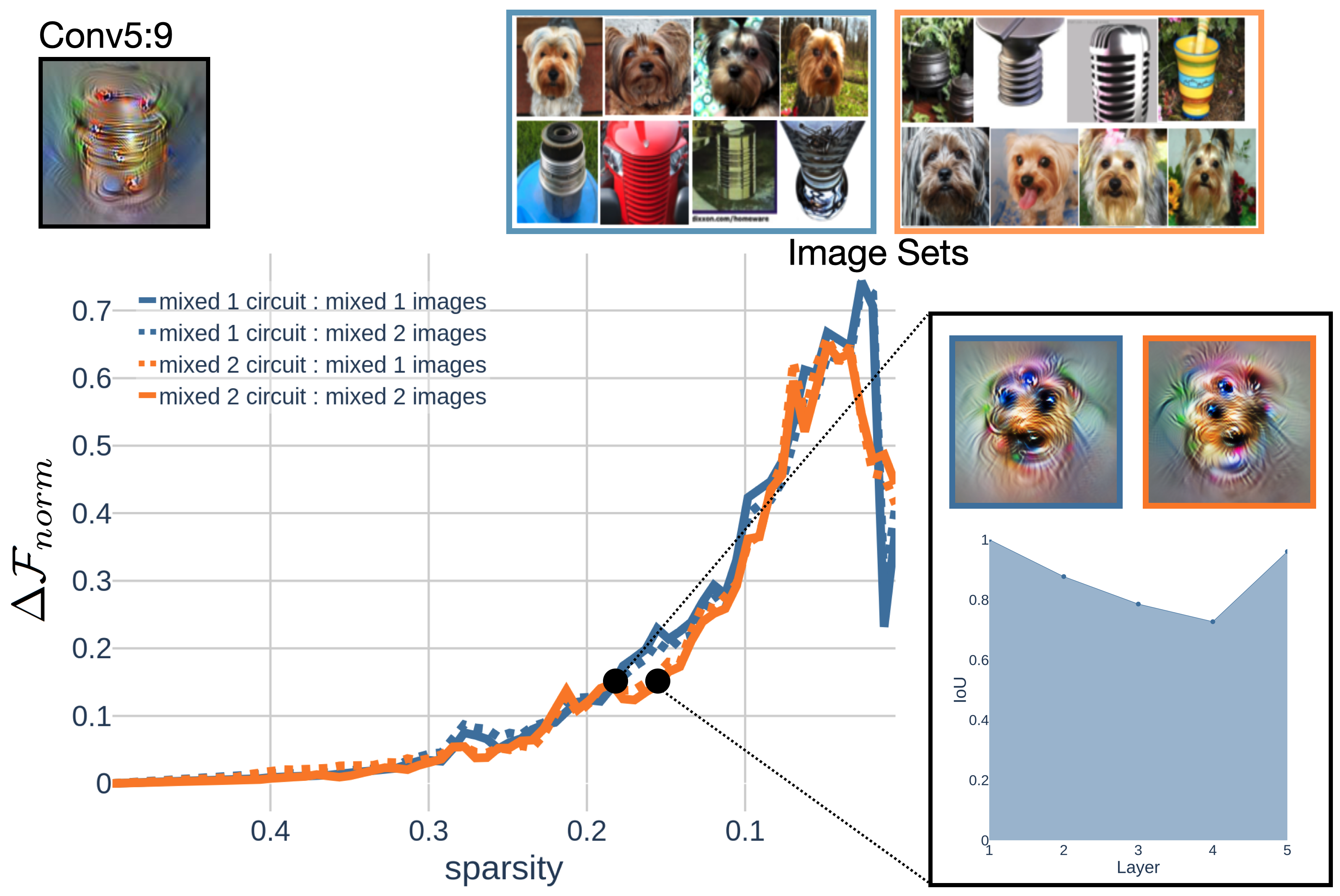} 
        \caption{subcircuits computed from mixed cluster}\label{fig:poly mixed}
    \end{minipage}

\end{figure*}

\section{Circuit Diagrams}\label{sec:diagram}

The above results demonstrate that sparse circuits exist in CNNs that well approximate the responses of targeted features. In some cases, these approximate circuits are so sparse that their full computation graph can be rendered in a single diagram, without excluding any weights. This allows for a detailed, mechanistic inspection of how latent features are built. To facilitate this inspection, we have developed a tool for \textit{extended circuit diagrams}. Extended circuit diagrams follow conventions developed in Olah et al (2020)\cite{circuitszoomin}, where our circuit pruning allows us to extend these diagrams hierarchically all the way back to pixel space. 

For illustration, Figure \ref{fig:small_circuit_diagram} shows an extended circuit diagram for \textit{conv3:56} (the 56th filter in the 3rd convolutional layer) in Alexnet. Where \textit{conv3:56} had 12864 preceding kernels in the original model, we've extracted an approximate circuit with highly correlated activations (see inset scatter plot) that can be diagrammed with only 13 kernels. In this diagram, vertices represent filters, and edges represent convolution with a 2D kernel. The width of each edge is proportional to the \textit{actgrad} salience score it received when pruning the circuit, and its color codes the average sign of the kernel's weights. Feature visualizations are displayed over their corresponding vertex, rendered with respect to the circuit, \textit{not the original model} (unlike \cite{circuitszoomin}). This ensures the visualization is representative of the filtering operations displayed in the diagram. Inspecting this diagram, following the thickest edges reveals the most critical computation for \textit{conv3:56}, which apparently checks for the presence of a magenta dot in the middle right of its receptive field. Given the original feature visualization alone (top right), one might presume a far more complex function underlies this feature. \par

\section{Subcircuit Pruning}\label{sec:subcircuits}

The circuits we've extracted so far approximate features as a whole, preserving their activations to a broad sample of images. However, in many cases it may be possible to further decompose a feature into even sparser \textit{subcircuits}, each responsible for computing a feature's response to a specific image. As our saliency-based circuit pruning method is image conditional, it should be well-suited to identify such subcircuits. In the following experiments we demonstrate two cases of hypothesis-driven subcircuit pruning, dissecting polysemantic features across their categories, and dissecting a circle detector into subcircuits for simpler parts.

\subsection{Polysemanticity}\label{sec:polysemantic}

Polysemantic filters are those that activate highly and selectively for images of seemingly disparate semantic categories. Many instances of such filters have been observed in CNNs \cite{polysemanticandreas,olah2017feature,circuitszoomin,polysemanticvec}. What computations account for these activations to different categories? For the sake of concreteness, suppose we have identified a polysemantic filter that responds highly to cats and cars. It could be the case that there is actually a shared feature common to images of cats and cars that underlies the filter's high activations, a feature which is hidden to our visual introspections because we are overwhelmed by the semantic difference of the two categories. If this were the case, it should not be possible to isolate the computation responsible for the filter's response to only cats or cars, as the computation is the same for both categories, i.e. the detection of the shared feature. Alternatively, it has been proposed that polysemanticity is best understood as a superficial entanglement of distinct features enabling network compression \cite{superposition}. In this case, there should exist distinct computations in the network responsible for the filter's response to cats and cars. Subcircuit pruning enables us to distinguish between these possibilities empirically.
\par
Before we could attempt to extract subcircuits from poly-semantic filters, we needed a way of identifying them. Rather than search through the top activating images across hundreds of filters in Alexnet, we took a data-driven approach. We reasoned that a polysemantic filter should return highest activations to a set of images that have clusterable hidden vectors in the feature's layer. Selecting for filters with this property identifies many with intuitive polysemanticity (when one views the images they respond most highly to). For a full treatment of this approach, see the appendix.

\par For this example, we focus on \textit{Conv5:9} in regularized Alexnet, identified using our clustering procedure. Qualitatively, the image patches corresponding to this filter's highest activations (cropped to individual activation's receptive field \cite{recep_field_1}) look to belong to the distinct categories of terrier faces and metal cylinders with horizontal lines. Example image patches from each category can be seen in the top right of Figure \ref{fig:poly clusters}. Let \(\mathcal{D}_{\text{dog}}\) refer to the set of exhitatory terrier images, then we can prune for the subcircuit responsible for \textit{Conv5:9}'s response to terriers using the saliency criteria \(\boldsymbol{S}(\boldsymbol{\theta};\textit{Conv5:9},\mathcal{D}_{\text{dog}})\). Conversely, we can prune for the metal cylinder detecting subcircuit with the criteria \(\boldsymbol{S}(\boldsymbol{\theta};\textit{Conv5:9},\mathcal{D}_{\text{metal}})\). In contrast to section \ref{sec:methodcompare}, where gradients in the saliency computation were with respect to the summed activation map per image, \(\sum|\mathbf{A}_{f}|\), here the gradients are with respect to the top activation in each map. These are the particular activations we are attempting to preserve. Using the distinct criteria for terriers and metal cylinders, we pruned subcircuits at 70 linearly spaced sparsities, preserving 50\%-.5\% of the relevant kernels in the original model. To quantify subcircuit feature preservation, Pearson correlation is no longer an appropriate metric, given the small, homongenous set of activations under consideration. Instead, we simply measure the absolute difference between the targeted activations in the original model and those in a circuit.  We normalize this measure by the mean of the original activations. 
 
With this metric, lower numbers show better preservation of the original activations. The solid lines in Figure\ref{fig:poly clusters} show how the activations the circuit was meant to preserve deviate from the original activations (blue for the terrier-preserving circuit, orange for the metal cylinder preserving-circuit). Overall, we were able to extract  subcircuits that preserve their target activations to high levels of sparsity. 
 \par 
 Critically, we next examined how well these subcircuits responded to the other set of image patches (dotted lines). If \textit{Conv5:9} computes its response to terriers and metal lines by distinct means, then these activations should be far less preserved. Conversely, if \textit{Conv5:9} activates for terriers and metal lines due to a common feature between the categories, a subcircuit that preserves one category's activations should also preserve the other's. In such a case, the dotted lines would fall on top of the solid lines. The dotted lines in this plot clearly deviate at lower sparsities than the solid lines; thus, we've identified subcircuits that compute \textit{Conv5:9} responses to only terriers but not metal cylinders, and visa-versa. \par
 How overlapping are these category-wise subcircuits? To answer this question, we computed the intersection over union (IoU) of the sets of kernels belonging to each circuit. Specifically, we compared the circuits at the last sparsity for which our \(\Delta\ f_{norm}\) metric remains below .15 for each circuits' target activations. This intersection over union is shown as a function of layer depth in the bottom-right of Figure\ref{fig:small_circuit_diagram}. It seems the two subcircuits utilize the same layer 1 computations (e.g. gabors, center-surround filters etc.) but deviate from one another in subsequent layers. We also render feature visualizations for these two subcircuits, shown above the IoU plot. These show a clear distinction between the subcircuits as well. See the appendix for an identical analysis performed on another polysemantic feature, which detects both monkey faces and written-text.
 \par 
As a control analysis, we extracted two new subcircuits, but this time with respect to a random 50/50 split of image patches from \(\mathcal{D}_{dog}\) and \(\mathcal{D}_{metal}\). We hypothesized that these subcircuits would not be separable, as the subcircuit that preserves activations for one half of the metal cylinders and terriers images would be identical to the subcircuit for the other halves. The results are shown in Figure\ref{fig:poly mixed} and confirm our hypothesis, indicated by the overlap between the solid and dashed lines, and the high IoU plot across the layers.  
  \par

\subsection{Circle Parts}\label{sec:circle}
While one might expect a polysemantic feature can be decomposed into category-wise subcircuits, is this the only case, or might subcircuit decomposition apply to features that don't have clear polysemanticity? Our next experiment applies to a different filter, (\textit{conv4:53} in Alexnet), which appears to be a concentric circle detector, given its feature visualization and top activating dataset examples (Figure\ref{fig:concentric_circles}). It is hypothesized that in the general case, shape detectors can be constructed by the composition of simpler curve/line detectors. Is that the case for this circle detector, and further -- are distinct curves needed to account for its activation to circles at different scales? \par

\begin{figure}[h!]
  \centering

   \includegraphics[width=0.8\linewidth]{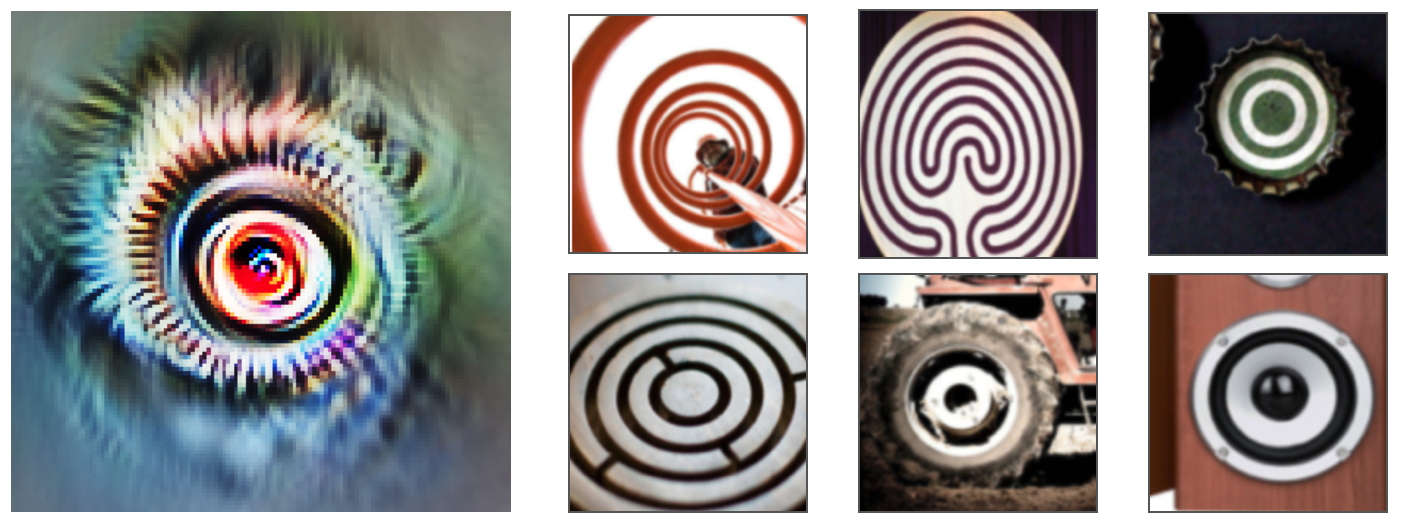}

   \caption{Concentric circle detector.}
   \label{fig:concentric_circles}
\end{figure}

\begin{figure*}
  \centering
  \includegraphics[width=\textwidth]{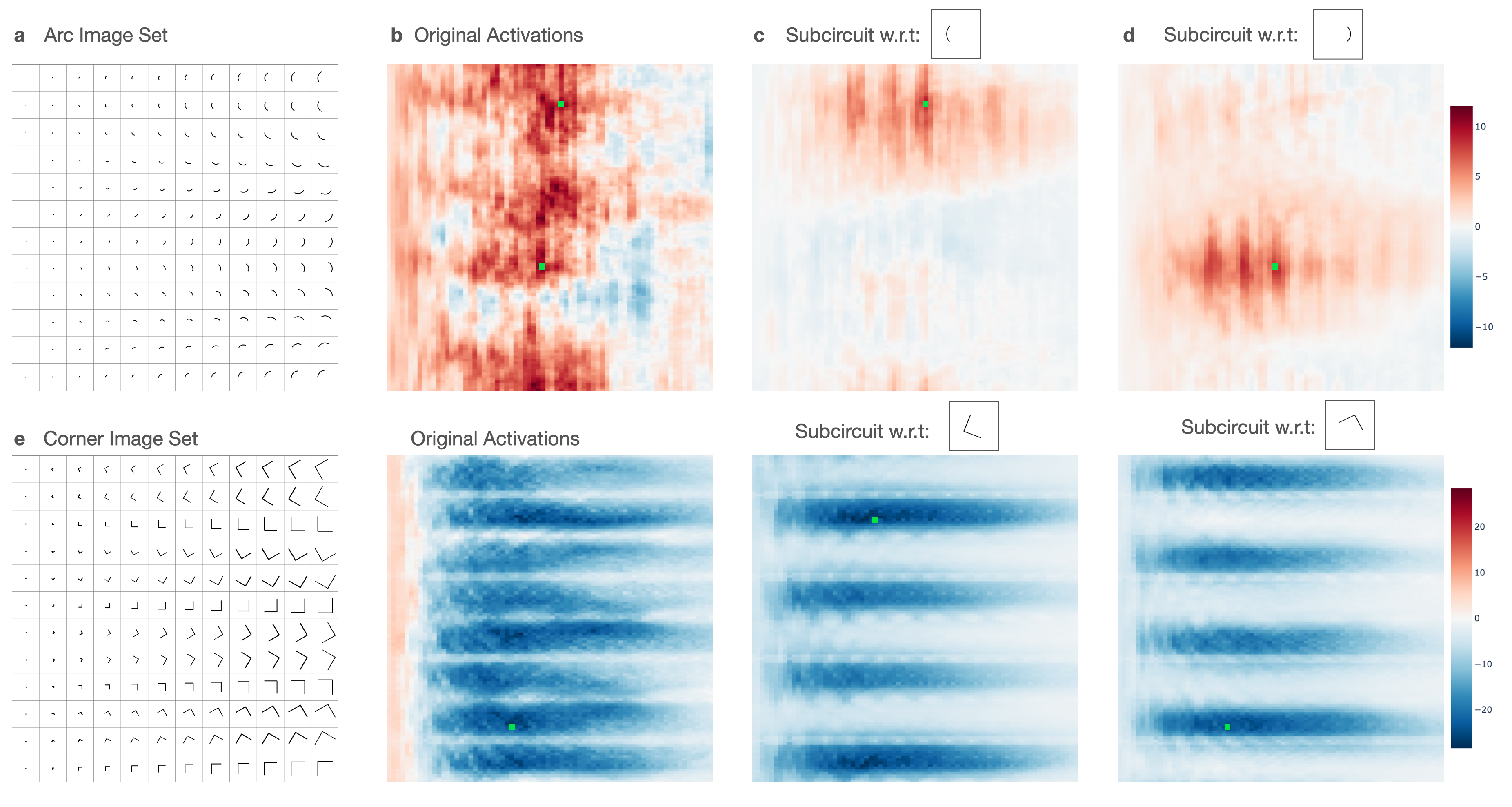}

  \caption{Pruning feature \textit{conv4:53} into image-wise subcircuits. Green dots mark images used to prune subcircuits.}\label{fig:circle_param}
\end{figure*}
As one way to test these compositional hypotheses, we created a probe set of images depicting circular arcs of 90 degrees, parameterized in two dimensions by the radius of the circle and the angle to the arc's midpoint (Figure \ref{fig:circle_param}a}). We fit these images to the receptive field of the central position in \textit{conv4:53}'s activation map, then compute the activations at this position in response to all the images, spanning the rotation/size parameter space. Figure \ref{fig:circle_param}b shows a heatmap of these activations plotted in the parameter space, revealing broadly positive activations to arcs, and a frequency pattern in both the rotation and size dimensions. \par
Are different areas on this parameterized activation surface attributable to separable subcircuits? To address this question, we chose a high 'peak' on the surface and pruned a subcircuit for the corresponding image/activation, following the procedure in section \ref{sec:polysemantic}. We then recomputed the activation surface for this pruned circuit, to assess what was preserved relative to the original feature's surface. Figure \ref{fig:circle_param}c reveals the subcircuit has a preserved response in the size dimension, but has lost sensitivity to arcs at rotations different from its target arc. This result suggests there exists separable subcircuits with distinct weights that compute \textit{conv4:53}'s response to arcs at different rotations, but a shared subcircuit for computing its response to arcs of the same rotation, but at different scales. To verify this observation, we pruned a second subcircuit with respect to a peak activation at a different arc rotation. The activation surface for this subcircuit (Figure\ref{fig:circle_param}.d) shows a symmetrical result to the first, preserving scale, but not the the first subcircuit's rotations. Whats more these subcircuits have a very low IoU, with \(<.1\) shared kernels across all layers.\par
We next took one further hypothesis driven step into this feature decomposition, by probing how this feature responds to right angle corners rather than arcs, again parameterized by size and rotation.  Corners might partially activate arc detectors, as a corner is tangent to the identically parameterized arc at its end points. However, we reasoned that the feature may incorporate designated subcircuits that inhibit its response to corners, making for a 'circle selective' feature, rather than a 'circle or square' feature. To test this hypothesis, we created a second probe set matched to the first but with right angle corners instead of arcs \ref{fig:circle_param}e. We see that indeed, activations for corners are negative across the parameter space, with a frequency pattern in the rotational dimension, and a flat response in the scale dimension. Further, the subcircuits pruned with respect to selected images again show preserved inhibition across scale, with separable circuits for different corner angles (mod \(45^{\circ}\)). As before, these two corner-wise subcircuits have a very low IoU\(<.1\). We hypothesize that these results can be attributed to shift invariances induced by convolution (see appendix).
\section{Conclusion}\label{sec:conclusion}
In this work we introduce feature-preserving circuit pruning, a novel approach to network modularization. We show that the function that computes a latent filters activations from pixels  can be well-approximated by sparse paths of computation that hiearchically route through the network. We compare different pruning saliency criteria towards this end, finding they are similarly effective for identifying sparse circuits, and better than magnitude-based pruning. We further demonstrate how sparse sub-circuits can be extracted that preserve features' responses to only a selection of images, providing mechanistic insight for two case studies (polysemanticity and concentric circle detector).  We also provide a general circuit pruning API and a circuit diagramming tool. While here we have focused on the structured pruning of convolutional kernels, our method is general and can be applied to the pruning of individual weights or entire convolutional filters (available through our provided circuit pruning API). Similarly, while we have focused our experiments on circuits that compute the activations of convolutional filters, our technique accommodates a much broader notion of feature preservation. In the future, we hope to extend this work to other model architectures (transformers) trained on a diversity of objectives.


\bibliographystyle{ieee_fullname}
\bibliography{egbib}

\begin{thebibliography}{10}\itemsep=-1pt

\bibitem{neural_module_networks}
Jacob Andreas, Marcus Rohrbach, Trevor Darrell, and Dan Klein.
\newblock Neural module networks.
\newblock In {\em Proceedings of the IEEE conference on computer vision and
  pattern recognition}, pages 39--48, 2016.

\bibitem{visual_question_answering}
Stanislaw Antol, Aishwarya Agrawal, Jiasen Lu, Margaret Mitchell, Dhruv Batra,
  C~Lawrence Zitnick, and Devi Parikh.
\newblock Vqa: Visual question answering.
\newblock In {\em Proceedings of the IEEE international conference on computer
  vision}, pages 2425--2433, 2015.

\bibitem{datasetintrinsic}
Sean Bell, Kavita Bala, and Noah Snavely.
\newblock Intrinsic images in the wild.
\newblock {\em ACM Transactions on Graphics (TOG)}, 33(4):1--12, 2014.

\bibitem{conditional_computation}
Emmanuel Bengio, Pierre-Luc Pierre-Luc~Bacon, Joelle Pineau, and Doina Precup.
\newblock Conditional computation in neural networks for faster models.
\newblock In {\em ICLR Workshop}. International Conference on Learning
  Representations, ICLR, 2018.

\bibitem{BLAS}
L~Susan Blackford, Antoine Petitet, Roldan Pozo, Karin Remington, R~Clint
  Whaley, James Demmel, Jack Dongarra, Iain Duff, Sven Hammarling, Greg Henry,
  et~al.
\newblock An updated set of basic linear algebra subprograms (blas).
\newblock {\em ACM Transactions on Mathematical Software}, 28(2):135--151,
  2002.

\bibitem{circuitsthread}
Nick Cammarata, Shan Carter, Gabiel Goh, Chris Olah, Michael Petrov, Ludwig
  Schubert, Chelsea Voss, Ben Egan, and Swee~Kiat Lim.
\newblock Thread: Circuits.
\newblock {\em Distill}, 2020.
\newblock https://distill.pub/2020/circuits.

\bibitem{circuit_curve}
Nick Cammarata, Gabriel Goh, Shan Carter, Chelsea Voss, Ludwig Schubert, and
  Chris Olah.
\newblock Curve circuits.
\newblock {\em Distill}, 2021.
\newblock https://distill.pub/2020/circuits/curve-circuits.

\bibitem{dino}
Mathilde Caron, Hugo Touvron, Ishan Misra, Herv{\'e} J{\'e}gou, Julien Mairal,
  Piotr Bojanowski, and Armand Joulin.
\newblock Emerging properties in self-supervised vision transformers.
\newblock In {\em Proceedings of the IEEE/CVF International Conference on
  Computer Vision}, pages 9650--9660, 2021.

\bibitem{simCLR}
Ting Chen, Simon Kornblith, Mohammad Norouzi, and Geoffrey Hinton.
\newblock A simple framework for contrastive learning of visual
  representations.
\newblock In {\em International conference on machine learning}, pages
  1597--1607. PMLR, 2020.

\bibitem{datasetparts}
Xianjie Chen, Roozbeh Mottaghi, Xiaobai Liu, Sanja Fidler, Raquel Urtasun, and
  Alan Yuille.
\newblock Detect what you can: Detecting and representing objects using
  holistic models and body parts.
\newblock In {\em Proceedings of the IEEE conference on computer vision and
  pattern recognition}, pages 1971--1978, 2014.

\bibitem{datasettextures}
Mircea Cimpoi, Subhransu Maji, Iasonas Kokkinos, Sammy Mohamed, and Andrea
  Vedaldi.
\newblock Describing textures in the wild.
\newblock In {\em Proceedings of the IEEE conference on computer vision and
  pattern recognition}, pages 3606--3613, 2014.

\bibitem{branch_multicolumn}
Dan Ciregan, Ueli Meier, and Jürgen Schmidhuber.
\newblock Multi-column deep neural networks for image classification.
\newblock In {\em 2012 IEEE Conference on Computer Vision and Pattern
  Recognition}, pages 3642--3649, 2012.

\bibitem{modules_by_weight_mask}
R{\'o}bert Csord{\'a}s, Sjoerd van Steenkiste, and J{\"u}rgen Schmidhuber.
\newblock Are neural nets modular? inspecting functional modularity through
  differentiable weight masks.
\newblock {\em arXiv preprint arXiv:2010.02066}, 2020.

\bibitem{FORCE}
Pau de Jorge, Amartya Sanyal, Harkirat Behl, Philip Torr, Gr{\'e}gory Rogez,
  and Puneet~K Dokania.
\newblock Progressive skeletonization: Trimming more fat from a network at
  initialization.
\newblock In {\em International Conference on Learning Representations}, 2020.

\bibitem{imagenet_cvpr09}
J. Deng, W. Dong, R. Socher, L.-J. Li, K. Li, and L. Fei-Fei.
\newblock {ImageNet: A Large-Scale Hierarchical Image Database}.
\newblock In {\em CVPR09}, 2009.

\bibitem{oraclefilter}
Xiaohan Ding, Guiguang Ding, Yuchen Guo, Jungong Han, and Chenggang Yan.
\newblock Approximated oracle filter pruning for destructive cnn width
  optimization.
\newblock In {\em International Conference on Machine Learning}, pages
  1607--1616. PMLR, 2019.

\bibitem{superposition}
Nelson Elhage, Tristan Hume, Catherine Olsson, Nicholas Schiefer, Tom Henighan,
  Shauna Kravec, Zac Hatfield-Dodds, Robert Lasenby, Dawn Drain, Carol Chen,
  Roger Grosse, Sam McCandlish, Jared Kaplan, Dario Amodei, Martin Wattenberg,
  and Christopher Olah.
\newblock Toy models of superposition.
\newblock {\em Transformer Circuits Thread}, 2022.
\newblock https://transformer-circuits.pub/2022/toy\_model/index.html.

\bibitem{deepviz}
Dumitru Erhan, Yoshua Bengio, Aaron Courville, and Pascal Vincent.
\newblock Visualizing higher-layer features of a deep network.
\newblock {\em Technical Report, Univeristé de Montréal}, 01 2009.

\bibitem{self-supervised-transfer}
Linus Ericsson, Henry Gouk, and Timothy~M Hospedales.
\newblock How well do self-supervised models transfer?
\newblock In {\em Proceedings of the IEEE/CVF Conference on Computer Vision and
  Pattern Recognition}, pages 5414--5423, 2021.

\bibitem{spectral_clustering_1}
Daniel Filan, Stephen Casper, Shlomi Hod, Cody Wild, Andrew Critch, and Stuart
  Russell.
\newblock Clusterability in neural networks.
\newblock {\em arXiv preprint arXiv:2103.03386}, 2021.

\bibitem{polysemanticvec}
Ruth Fong and Andrea Vedaldi.
\newblock Net2vec: Quantifying and explaining how concepts are encoded by
  filters in deep neural networks.
\newblock In {\em Proceedings of the IEEE conference on computer vision and
  pattern recognition}, pages 8730--8738, 2018.

\bibitem{filteredge}
Chinthaka Gamanayake, Lahiru Jayasinghe, Benny Kai~Kiat Ng, and Chau Yuen.
\newblock Cluster pruning: An efficient filter pruning method for edge ai
  vision applications.
\newblock {\em IEEE Journal of Selected Topics in Signal Processing},
  14(4):802--816, 2020.

\bibitem{gilpin2019explaining}
Leilani~H. Gilpin, David Bau, Ben~Z. Yuan, Ayesha Bajwa, Michael Specter, and
  Lalana Kagal.
\newblock Explaining explanations: An overview of interpretability of machine
  learning.
\newblock {\em International Conference on Data Science and Advanced
  Analytics}, 2019.

\bibitem{resnet}
Kaiming He, Xiangyu Zhang, Shaoqing Ren, and Jian Sun.
\newblock Deep residual learning for image recognition.
\newblock In {\em Proceedings of the IEEE conference on computer vision and
  pattern recognition}, pages 770--778, 2016.

\bibitem{softfilter}
Yang He, Guoliang Kang, Xuanyi Dong, Yanwei Fu, and Yi Yang.
\newblock Soft filter pruning for accelerating deep convolutional neural
  networks.
\newblock {\em arXiv preprint arXiv:1808.06866}, 2018.

\bibitem{spectral_clustering_2}
Shlomi Hod, Daniel Filan, Stephen Casper, Andrew Critch, and Stuart Russell.
\newblock Quantifying local specialization in deep neural networks.
\newblock {\em arXiv e-prints}, pages arXiv--2110, 2021.

\bibitem{neural_module_explain}
Ronghang Hu, Jacob Andreas, Trevor Darrell, and Kate Saenko.
\newblock Explainable neural computation via stack neural module networks.
\newblock In {\em Proceedings of the European conference on computer vision
  (ECCV)}, pages 53--69, 2018.

\bibitem{data_routing_paths_2}
Ashkan Khakzar, Soroosh Baselizadeh, Saurabh Khanduja, Christian Rupprecht,
  Seong~Tae Kim, and Nassir Navab.
\newblock Neural response interpretation through the lens of critical pathways.
\newblock In {\em Proceedings of the IEEE/CVF Conference on Computer Vision and
  Pattern Recognition}, pages 13528--13538, 2021.

\bibitem{lucent}
Lim~Swee Kiat.
\newblock Lucent, 2020.
\newblock https://github.com/greentfrapp/lucent.

\bibitem{module_learning}
Louis Kirsch, Julius Kunze, and David Barber.
\newblock Modular networks: Learning to decompose neural computation.
\newblock {\em Advances in neural information processing systems}, 31, 2018.

\bibitem{alexnet}
Alex Krizhevsky, Ilya Sutskever, and Geoffrey Hinton.
\newblock Imagenet classification with deep convolutional neural networks.
\newblock {\em Neural Information Processing Systems}, 25, 01 2012.

\bibitem{recep_field_1}
Hung Le and Ali Borji.
\newblock What are the receptive, effective receptive, and projective fields of
  neurons in convolutional neural networks?
\newblock {\em CoRR}, abs/1705.07049, 2017.

\bibitem{snip}
Namhoon Lee, Thalaiyasingam Ajanthan, and Philip Torr.
\newblock Snip: Single-shot network pruning based on connection sensitivity.
\newblock In {\em International Conference on Learning Representations}, 2018.

\bibitem{kernel_pruning_interpret}
Yuchao Li, Shaohui Lin, Baochang Zhang, Jianzhuang Liu, David Doermann,
  Yongjian Wu, Feiyue Huang, and Rongrong Ji.
\newblock Exploiting kernel sparsity and entropy for interpretable cnn
  compression.
\newblock In {\em Proceedings of the IEEE/CVF Conference on Computer Vision and
  Pattern Recognition (CVPR)}, June 2019.

\bibitem{kernel_regularization}
Chen Lin, Zhao Zhong, Wu Wei, and Junjie Yan.
\newblock Synaptic strength for convolutional neural network.
\newblock {\em Advances in Neural Information Processing Systems}, 31, 2018.

\bibitem{globalfilter}
Shaohui Lin, Rongrong Ji, Yuchao Li, Yongjian Wu, Feiyue Huang, and Baochang
  Zhang.
\newblock Accelerating convolutional networks via global \& dynamic filter
  pruning.
\newblock In {\em IJCAI}, volume~2, page~8, 2018.

\bibitem{ganpruning}
Shaohui Lin, Rongrong Ji, Chenqian Yan, Baochang Zhang, Liujuan Cao, Qixiang
  Ye, Feiyue Huang, and David Doermann.
\newblock Towards optimal structured cnn pruning via generative adversarial
  learning.
\newblock In {\em Proceedings of the IEEE/CVF Conference on Computer Vision and
  Pattern Recognition}, pages 2790--2799, 2019.

\bibitem{filterentropy}
Jian-Hao Luo and Jianxin Wu.
\newblock An entropy-based pruning method for cnn compression.
\newblock {\em arXiv preprint arXiv:1706.05791}, 2017.

\bibitem{autofilter}
Jian-Hao Luo and Jianxin Wu.
\newblock Autopruner: An end-to-end trainable filter pruning method for
  efficient deep model inference.
\newblock {\em Pattern Recognition}, 107:107461, 2020.

\bibitem{thinet}
Jian-Hao Luo, Hao Zhang, Hong-Yu Zhou, Chen-Wei Xie, Jianxin Wu, and Weiyao
  Lin.
\newblock Thinet: pruning cnn filters for a thinner net.
\newblock {\em IEEE transactions on pattern analysis and machine intelligence},
  41(10):2525--2538, 2018.

\bibitem{kernelexplore}
Huizi Mao, Song Han, Jeff Pool, Wenshuo Li, Xingyu Liu, Yu Wang, and William~J
  Dally.
\newblock Exploring the regularity of sparse structure in convolutional neural
  networks.
\newblock {\em arXiv preprint arXiv:1705.08922}, 2017.

\bibitem{HDBScan}
Leland McInnes, John Healy, and Steve Astels.
\newblock hdbscan: Hierarchical density based clustering.
\newblock {\em Journal of Open Source Software}, 2(11):205, 2017.

\bibitem{hierarch_reg}
Kakeru Mitsuno, Junichi Miyao, and Takio Kurita.
\newblock Hierarchical group sparse regularization for deep convolutional
  neural networks.
\newblock pages 1--8, 07 2020.

\bibitem{actgrad}
P Molchanov, S Tyree, T Karras, T Aila, and J Kautz.
\newblock Pruning convolutional neural networks for resource efficient
  inference.
\newblock In {\em 5th International Conference on Learning Representations,
  ICLR 2017-Conference Track Proceedings}, 2019.

\bibitem{datasetcontext}
Roozbeh Mottaghi, Xianjie Chen, Xiaobai Liu, Nam-Gyu Cho, Seong-Whan Lee, Sanja
  Fidler, Raquel Urtasun, and Alan Yuille.
\newblock The role of context for object detection and semantic segmentation in
  the wild.
\newblock In {\em Proceedings of the IEEE conference on computer vision and
  pattern recognition}, pages 891--898, 2014.

\bibitem{mozer_prune}
Michael~C Mozer and Paul Smolensky.
\newblock Skeletonization: A technique for trimming the fat from a network via
  relevance assessment.
\newblock In D. Touretzky, editor, {\em Advances in Neural Information
  Processing Systems}, volume~1. Morgan-Kaufmann, 1988.

\bibitem{polysemanticandreas}
Jesse Mu and Jacob Andreas.
\newblock Compositional explanations of neurons.
\newblock {\em Advances in Neural Information Processing Systems},
  33:17153--17163, 2020.

\bibitem{circuitszoomin}
Chris Olah, Nick Cammarata, Ludwig Schubert, Gabriel Goh, Michael Petrov, and
  Shan Carter.
\newblock Zoom in: An introduction to circuits.
\newblock {\em Distill}, 2020.
\newblock https://distill.pub/2020/circuits/zoom-in.

\bibitem{circuit_equivariance}
Chris Olah, Nick Cammarata, Chelsea Voss, Ludwig Schubert, and Gabriel Goh.
\newblock Naturally occurring equivariance in neural networks.
\newblock {\em Distill}, 2020.
\newblock https://distill.pub/2020/circuits/equivariance.

\bibitem{olah2017feature}
Chris Olah, Alexander Mordvintsev, and Ludwig Schubert.
\newblock Feature visualization.
\newblock {\em Distill}, 2017.
\newblock https://distill.pub/2017/feature-visualization.

\bibitem{modules_inout_1}
Rangeet Pan and Hridesh Rajan.
\newblock On decomposing a deep neural network into modules.
\newblock In {\em Proceedings of the 28th ACM Joint Meeting on European
  Software Engineering Conference and Symposium on the Foundations of Software
  Engineering}, pages 889--900, 2020.

\bibitem{modules_inout_2}
Rangeet Pan and Hridesh Rajan.
\newblock Decomposing convolutional neural networks into reusable and
  replaceable modules.
\newblock In {\em Proceedings of the 44th International Conference on Software
  Engineering}, pages 524--535, 2022.

\bibitem{branching_pc_dropout}
Kien~Tuong Phan, Tomas~Henrique Maul, Tuong~Thuy Vu, and Weng~Kin Lai.
\newblock Improving neural network generalization by combining parallel
  circuits with dropout.
\newblock In {\em Neural Information Processing: 23rd International Conference,
  ICONIP 2016, Kyoto, Japan, October 16--21, 2016, Proceedings, Part III 23},
  pages 572--580. Springer, 2016.

\bibitem{interpretability_survey}
Tilman R{\"a}ukur, Anson Ho, Stephen Casper, and Dylan Hadfield-Menell.
\newblock Toward transparent ai: A survey on interpreting the inner structures
  of deep neural networks.
\newblock {\em arXiv preprint arXiv:2207.13243}, 2022.

\bibitem{moe_routing}
Carlos Riquelme, Joan Puigcerver, Basil Mustafa, Maxim Neumann, Rodolphe
  Jenatton, Andr{\'e} Susano~Pinto, Daniel Keysers, and Neil Houlsby.
\newblock Scaling vision with sparse mixture of experts.
\newblock {\em Advances in Neural Information Processing Systems},
  34:8583--8595, 2021.

\bibitem{referential_expression_grounding}
Anna Rohrbach, Marcus Rohrbach, Ronghang Hu, Trevor Darrell, and Bernt Schiele.
\newblock Grounding of textual phrases in images by reconstruction.
\newblock In {\em Computer Vision--ECCV 2016: 14th European Conference,
  Amsterdam, The Netherlands, October 11--14, 2016, Proceedings, Part I 14},
  pages 817--834. Springer, 2016.

\bibitem{routingnetworks}
Clemens Rosenbaum, Tim Klinger, and Matthew Riemer.
\newblock Routing networks: Adaptive selection of non-linear functions for
  multi-task learning.
\newblock In {\em International Conference on Learning Representations}.
  International Conference on Learning Representations, ICLR, 2018.

\bibitem{vgg}
Karen Simonyan and Andrew Zisserman.
\newblock Very deep convolutional networks for large-scale image recognition.
\newblock {\em arXiv preprint arXiv:1409.1556}, 2014.

\bibitem{inception}
Christian Szegedy, Wei Liu, Yangqing Jia, Pierre Sermanet, Scott Reed, Dragomir
  Anguelov, Dumitru Erhan, Vincent Vanhoucke, and Andrew Rabinovich.
\newblock Going deeper with convolutions.
\newblock In {\em Proceedings of the IEEE conference on computer vision and
  pattern recognition}, pages 1--9, 2015.

\bibitem{snipit}
Stijn Verdenius, Maarten Stol, and Patrick Forr{\'{e}}.
\newblock Pruning via iterative ranking of sensitivity statistics.
\newblock {\em CoRR}, abs/2006.00896, 2020.

\bibitem{circuit_weightviz}
Chelsea Voss, Nick Cammarata, Gabriel Goh, Michael Petrov, Ludwig Schubert, Ben
  Egan, Swee~Kiat Lim, and Chris Olah.
\newblock Visualizing weights.
\newblock {\em Distill}, 2021.
\newblock https://distill.pub/2020/circuits/visualizing-weights.

\bibitem{grasp}
Chaoqi Wang, Guodong Zhang, and Roger Grosse.
\newblock Picking winning tickets before training by preserving gradient flow.
\newblock In {\em International Conference on Learning Representations}, 2020.

\bibitem{branch_wang2015multi}
Mingming Wang.
\newblock Multi-path convolutional neural networks for complex image
  classification.
\newblock {\em arXiv preprint arXiv:1506.04701}, 2015.

\bibitem{data_routing_paths_1}
Yulong Wang, Hang Su, Bo Zhang, and Xiaolin Hu.
\newblock Interpret neural networks by identifying critical data routing paths.
\newblock In {\em proceedings of the IEEE conference on computer vision and
  pattern recognition}, pages 8906--8914, 2018.

\bibitem{watanabe2019interpreting}
Chihiro Watanabe.
\newblock Interpreting layered neural networks via hierarchical modular
  representation.
\newblock In {\em Neural Information Processing: 26th International Conference,
  ICONIP 2019, Sydney, NSW, Australia, December 12--15, 2019, Proceedings, Part
  V 26}, pages 376--388. Springer, 2019.

\bibitem{Yosinski_deepviz_toolbox}
Jason Yosinski, Jeff Clune, Anh~Mai Nguyen, Thomas~J. Fuchs, and Hod Lipson.
\newblock Understanding neural networks through deep visualization.
\newblock {\em ICML}, abs/1506.06579, 2015.

\bibitem{barlow-twins}
Jure Zbontar, Li Jing, Ishan Misra, Yann LeCun, and Stephane Deny.
\newblock Barlow twins: Self-supervised learning via redundancy reduction.
\newblock In Marina Meila and Tong Zhang, editors, {\em Proceedings of the 38th
  International Conference on Machine Learning}, volume 139 of {\em Proceedings
  of Machine Learning Research}, pages 12310--12320. PMLR, 18--24 Jul 2021.

\bibitem{netdissect}
Bolei Zhou, David Bau, Aude Oliva, and Antonio Torralba.
\newblock Interpreting deep visual representations via network dissection.
\newblock {\em IEEE transactions on pattern analysis and machine intelligence},
  41(9):2131--2145, 2018.

\bibitem{datasetade}
Bolei Zhou, Hang Zhao, Xavier Puig, Sanja Fidler, Adela Barriuso, and Antonio
  Torralba.
\newblock Scene parsing through ade20k dataset.
\newblock In {\em Proceedings of the IEEE conference on computer vision and
  pattern recognition}, pages 633--641, 2017.

\bibitem{kernel_pruning}
Jihong Zhu, Yang Zhao, and Jihong Pei.
\newblock Progressive kernel pruning based on the information mapping sparse
  index for cnn compression.
\newblock {\em IEEE Access}, 9:10974--10987, 2021.

\end{thebibliography}

\section{Appendix}
\subsection{Code}

 Code for this project is available at \footnote{https://github.com/chrishamblin7/circuit\textunderscore explorer/}{\href{https://github.com/chrishamblin7/circuit_explorer/}{this github repository}}.

\subsection{SNIP}\label{sec:SNIP}

While the SNIP saliency criterion was originally introduced in \cite{mozer_prune}, \cite{snip} demonstrated the criterion's efficacy for \textit{single-shot} pruning at initialization \cite{grasp,snipit,FORCE}, which utilizes no fine-tuning, making the criterion promising for our application. In SNIP, a binary mask \(\mathbf{c} \in\{0,1\}^{m} \) is inserted into the network, such that the new network is parameterized by \(\mathbf{c} \odot \boldsymbol{\theta}\), where \(\odot\) denotes the Hadamard product. In the original un-pruned network, \(\mathbf{c} = \{1\}^{m}\).  When \(c_{j} = 0\), this is equivalent to parameter \(\theta_{j}\) being pruned from the network. The authors reason the importance of parameter \(\theta_{j}\) to be proportional to the effect on the loss were it to be removed, or equivalently \(c_{j} = 0\):

\begin{equation}\label{eq:snip1}
\Delta \mathcal{L}_{j}(c_{j} ; \mathcal{D})=\mathcal{L}(c_{j} = 1; \mathcal{D})-\mathcal{L}\left(c_{j} = 0 ; \mathcal{D}\right), 
\end{equation}

 The authors' \textit{connection sensitivity} saliency criterion considers the gradient of \(\mathcal{L}\) with respect to \(c_{j}\). \(|\partial \mathcal{L} / \partial c_{j}|\) measures the sensitivity of \(\mathcal{L}\) to perturbations of \(c_{j}\), which they regard as an approximation of the \(|\Delta \mathcal{L}_{j}|\) induced by setting \(c_{j}=0\).
Their criterion can be equivalently calculated as the magnitude of parameters multiplied by their gradients:

\begin{equation}\label{eq:snip2}
\boldsymbol{S}_{snip}(\boldsymbol{\theta}):=\left|\partial \mathcal{L} / \partial \mathbf{c} \right|=\left|\frac{\partial \mathcal{L}(\boldsymbol{\theta} \odot \boldsymbol{c})}{\partial \boldsymbol{c}}\right|_{\mathbf{c}=\mathbf{1}}=\left|\frac{\partial \mathcal{L}}{\partial \boldsymbol{\theta}} \odot \boldsymbol{\theta}\right|
\end{equation}

\subsection{Force}\label{sec:FORCE}

The above SNIP criterion considers the effects of removing parameters in isolation, where in actuality we are endeavoring to remove many parameters from the network. Removal of a particular \(\theta_{j}\) may have a very different effect when the network is fully-intact versus removal when other parameters are also pruned. Suppose our sparse network is parameterized by \(\boldsymbol{\bar{\theta}}\), then \(\boldsymbol{S}_{snip}(\boldsymbol{\bar{\theta}})\) will certainly be different than \(\boldsymbol{S}_{snip}(\boldsymbol{\theta})\), as the gradients passing through the network will be different. Motivated by this observation, de Jorge et al (2020) \cite{FORCE} (and similarly \cite{snipit}) developed the \textit{FORCE} saliency criterion (foresight connection sensitivity), which attempts to identify important parameters for the resultant sparse network, despite the unfortunate circularity that this sparse network is supposed to be identified \textit{by way of} the saliency criterion. The author's make a subtle adjustment to the saliency-based pruning problem statement (eq \ref{eq:problem form saliency});

\begin{equation}\label{eq:force1}
\begin{gathered}
\underset{\boldsymbol{\bar{\theta}}}{\arg \max } \boldsymbol{S}_{force\_sum}(\boldsymbol{\bar{\theta}}) :=  \sum_{j\in supp(\boldsymbol{\bar{\theta}})}\boldsymbol{S}_{snip}(\bar{\theta_{j}}) \\
\text { s.t. } \quad \left\|\boldsymbol{\bar{\theta}}\right\|_{0} \leq \kappa , \quad \bar{\theta}_{j} \in \{\theta_{j},0\} \,\, \forall j \in \{1 \ldots m\}.
\end{gathered}
\end{equation}

They still endeavor to find a \(\boldsymbol{\bar{\theta}}\) with maximum cumulative saliency, but when parameter-wise saliency is measured in the pruned network, rather than the original network as in eq. \ref{eq:problem form saliency}. 
\par
Unfortunately eq. \ref{eq:force1} is not trivial to solve as is the original problem statement, but the authors reason an approximate solution can be obtained by iteratively masking an increasing number of parameters, recomputing  \(\boldsymbol{S}_{snip}(\boldsymbol{\bar{\theta}})\) at each step to get the next mask (by way of eq. 3). Let's call the network parameterization after the first masking iteration \(\boldsymbol{\bar{\theta}}_{1}\), the second \(\boldsymbol{\bar{\theta}}_{2}\), up to the final iteration \(T\) with a parameterization at the desired sparsity \(\boldsymbol{\bar{\theta}}_{T}\). The motivating intuition for this iterative sparsification follows from the observation that the fewer parameters are masked from \(\boldsymbol{\theta}\) to obtain \(\boldsymbol{\bar{\theta}}_{1}\), the better the \textit{gradient approximation} \(\frac{\delta\mathcal{L}(\boldsymbol{\theta})}{\delta\boldsymbol{\theta}} \approx \frac{\delta\mathcal{L}(\boldsymbol{\bar{\theta}}_{1})}{\delta\boldsymbol{\bar{\theta}}}_{1}\) should hold, and we can use \(\boldsymbol{S}_{snip}(\boldsymbol{\theta})\) to approximate \(\boldsymbol{S}_{snip}(\boldsymbol{\bar{\theta}}_{1})\). This logic is inductive; \(\boldsymbol{S}_{snip}(\boldsymbol{\bar{\theta}}_{1})\) is a better approximation of \(\boldsymbol{S}_{snip}(\boldsymbol{\bar{\theta}}_{2})\) than \(\boldsymbol{S}_{snip}(\boldsymbol{\theta})\) (as there are fewer parameters masked between \(\boldsymbol{\bar{\theta}}_{1}\) and \(\boldsymbol{\bar{\theta}}_{2}\)), \(\boldsymbol{S}_{snip}(\boldsymbol{\bar{\theta}}_{3})\) is better approximated by \(\boldsymbol{S}_{snip}(\boldsymbol{\bar{\theta}}_{2})\) than any of the previously computed scores, etc. In general \(S(\boldsymbol{\theta}_{T})\) should be calculated with many iterations of \(\boldsymbol{S}_{snip}\), resulting in a small difference in the masks between iterations, to optimize (eq \ref{eq:force1}). In this work we implement \textit{FORCE} with 10 iterations and the exponential decay scheduler from \cite{FORCE}, which were shown to be effective hyper-parameters in the original work. Given an iteration \(t\) and \(m\) relevant kernels, the number of kernels masked \(k_{t}\) at \(t\) is given by;

\begin{equation} \label{eq:force2}
k_{t}=\exp \{\alpha \log k+(1-\alpha) \log m\}, \quad \alpha=\frac{t}{T}
\end{equation}

We note that while optimizing eq. \ref{eq:force1} may lead to a \textit{trainable} model parameterization with high connection sensitivity (as was the intention in the original work), we explicitly want to minimize \(\Delta f(\boldsymbol{\theta},\boldsymbol{\theta}_{T})\). With \textit{FORCE}, at each iteration we are approximating \textit{an approximation}, \(\Delta f(\boldsymbol{\theta}_{i-1},\boldsymbol{\theta}_{i})\). This could lead to divergence from \(f\), depending on the accuracy of the iterative approximations (\cite{data_routing_paths_2} make a similar observation).
\begin{figure*}[t]
  \centering
  \includegraphics[width=\textwidth]{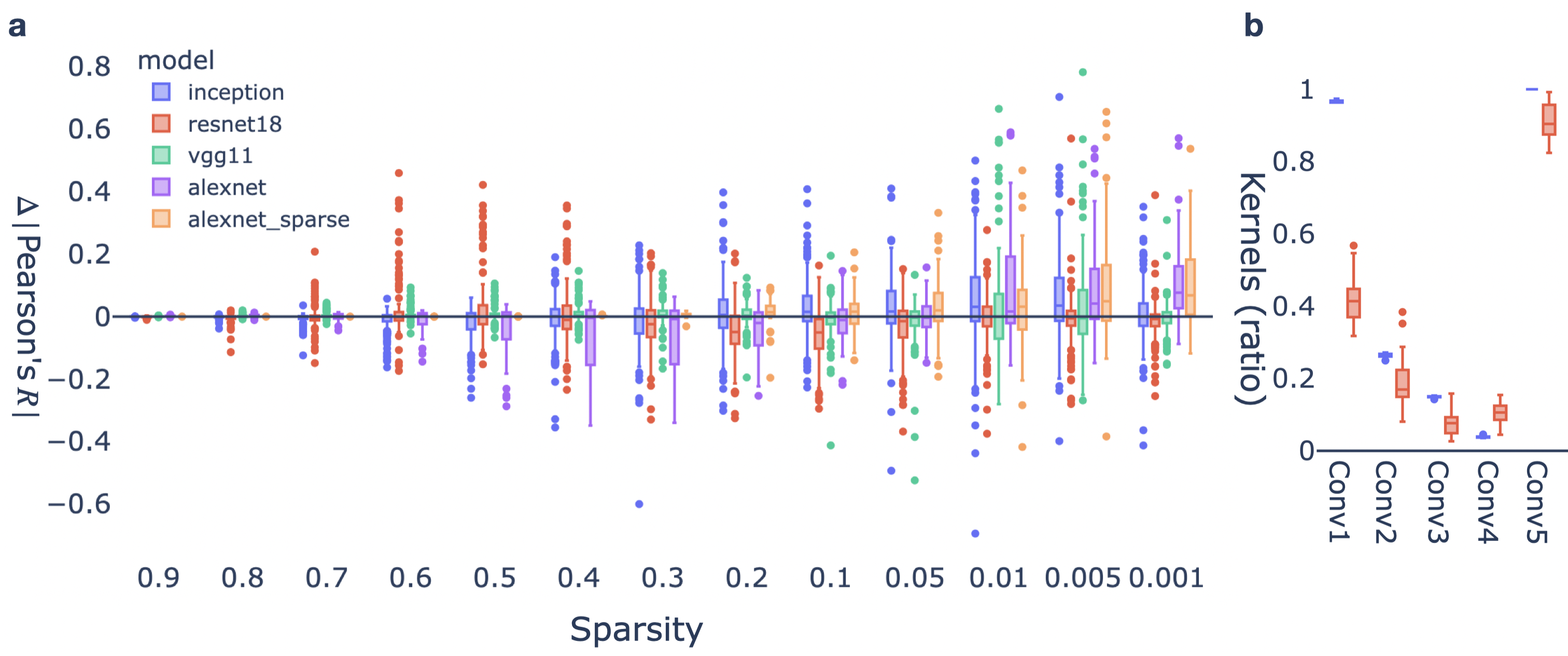}

  \caption{a) The effects of min-max normalizing scores per layer before pruning at different sparsities on feature preservation. The Y axis shows the difference in \(|\)Pearson's R\(|\)  when using min-max normalization (min-max - ~min-max). b) Circuits were pruned at .1 sparsity from sparsity-regularized Alexnet for 20 random features using kernel-wise \textit{actgrad} saliency criteria. For each convolutional layer (x-axis), we plot the ratio of relevant kernels kept in the circuit. Blue shows the circuit pruned with unnormalized scores, and red with min-max normalized scores.)}\label{fig:minmax}
\end{figure*}

\subsection{Circuits by Taylor Expansion}\label{sec:taylor}

Another way of deriving an approximation for \(\Delta \mathcal{L}\) given the removal of a parameter is by Taylor approximation. This approach was employed successfully by Molchanov et al (2019) \cite{actgrad} for the pruning of convolutional filters. Their saliency criterion concerns the activations in the network, rather than the weights themselves. Suppose some particular activation $a_{j}$ in the network were to be masked (set to 0), then the first order Taylor approximation of the loss is:

 \begin{equation} \label{eq:taylor1}
\mathcal{L}(a_{i}=0) = \mathcal{L}(a_{i}) - \frac{\delta \mathcal{L}}{\delta a_{i}}a_{i}
\end{equation}
 
As per usual, the authors wanted their salience criteria for an activation to approximate the magnitude of the change in loss when that activation is masked, which can be approximated with (eq \ref{eq:taylor1});

\begin{equation}\label{eq:taylor2}
 \boldsymbol{S}_{actgrad}(a_{i}):=\left| \mathcal{L}(a_{i}) - \mathcal{L}(a_{i}=0) \right| = \left|\frac{\partial \mathcal{L}}{\partial a_{i}} \odot a_{i}\right|
\end{equation}

Notice that this criterion is equivalent to \(\boldsymbol{S}_{snip}\), applied to activations as opposed to weights. Relatedly, the above Taylor derivation is applicable to \(\boldsymbol{S}_{snip}\), and as such we denote this saliency criteria \(\boldsymbol{S}_{actgrad}\), rather than \(\boldsymbol{S}_{taylor}\) as in \cite{actgrad}.

\begin{figure*}[t]
  \centering
  \includegraphics[width=\textwidth]{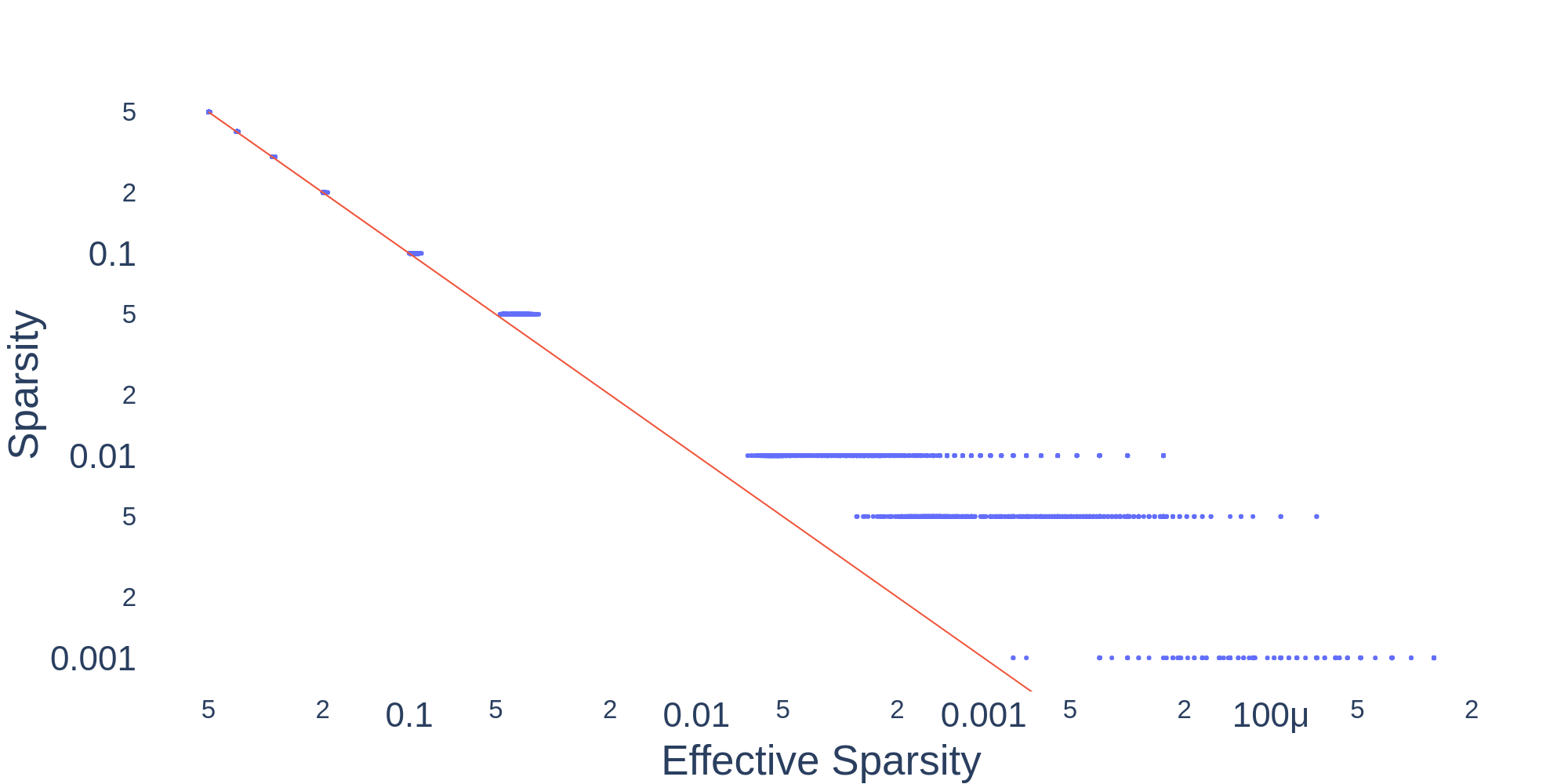}

  \caption{Sparsity versus effective sparsity of circuits}\label{fig:effective}
\end{figure*}
\begin{figure*}[t]
  \centering
  \includegraphics[width=\textwidth]{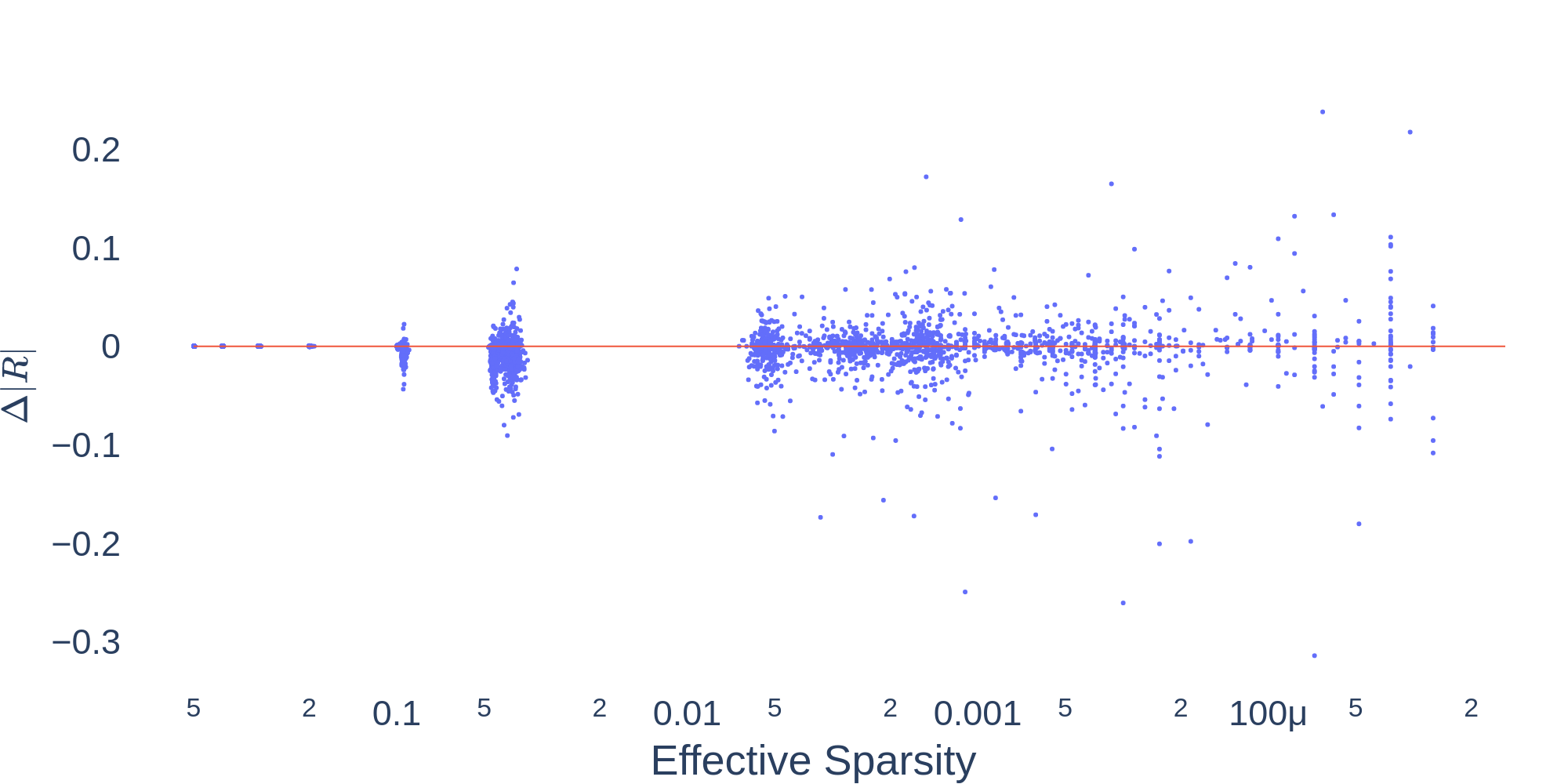}

  \caption{Circuit \(\Delta |R|\) (Pruned - Masked) }\label{fig:prunevmask}
\end{figure*}

\subsection{Min-max Normalization}
When pruning based on saliency scores, it is possible that parameters in a given layer yield systematically higher scores than others, and as a result the architecture has more parameters kept in a given layer than others. This could be beneficial if the optimal circuit architecture indeed favors a given layer, but it could also be a detrimental influence of exploding/vanishing gradients. The imbalance of saliency scores across layers can be negated by min-max normalizing the scores within each layer before pruning to \(\kappa\)-sparsity. Figure \ref{fig:minmax}.a show the difference in the difference in \(|\)Pearson's R\(|\) preservation metric yielded when applying this normalization to the \textit{ActGrad} saliency criteria across features/sparsities in different models. Depending on the feature, this normalization either improves or impairs feature preservation. Figure \ref{fig:minmax}.b shows this normalization indeed yields a different circuit architecture; for example, where the original circuits contained almost all the kernels in the first convolutional layer (blue), the min-max normed circuits only contains ~ 35-60\% of these kernels. 
\subsection{Circuit Disconnect}
At high sparsities a pruned circuit can become \textit{disconnected} from its target feature, meaning there no longer exists a connected path of kernel convolution operations from the input to the target feature. We found from our experiments pruning many circuits with the \textit{actgrad} criterion on regularized Alexnet, this happened only at the highest sparsities tested, .01, .005, and .001, with likelihoods of disconnection .057, .298, and .807 respectively. For the purposes of our reported Pearson R in the first experiments, we defined a disconnected circuit to have a Pearson R of 0.

\begin{figure*}[t]
  \centering
  \includegraphics[width=\textwidth]{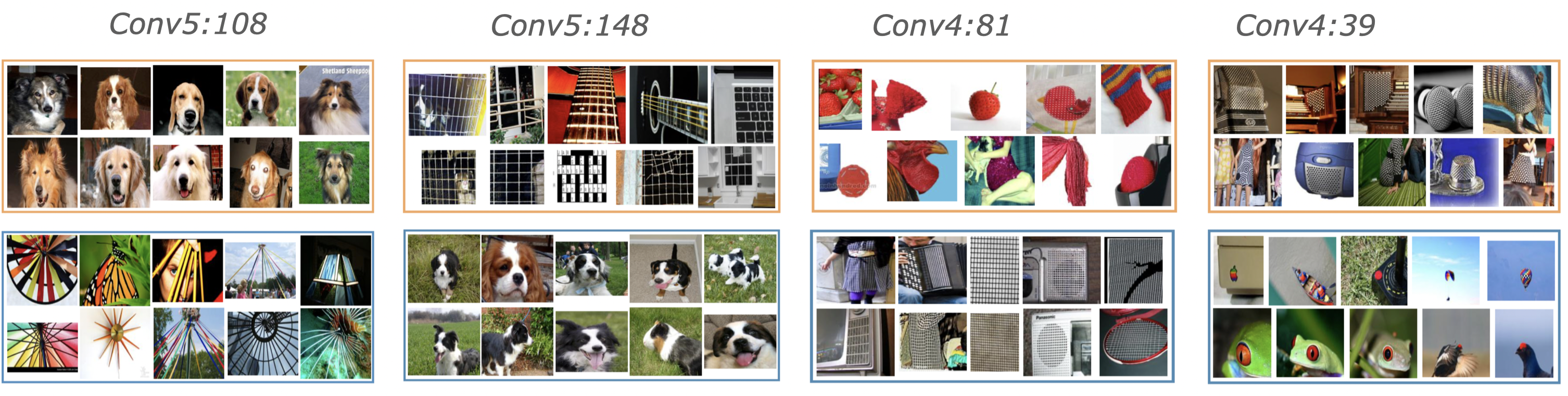}
  \caption{Examples of features with clusterable top activations (Orange and blue images belong to different clusters). }\label{fig:clusters}
\end{figure*}

\begin{figure*}[t]
  \centering
  \includegraphics[width=\textwidth]{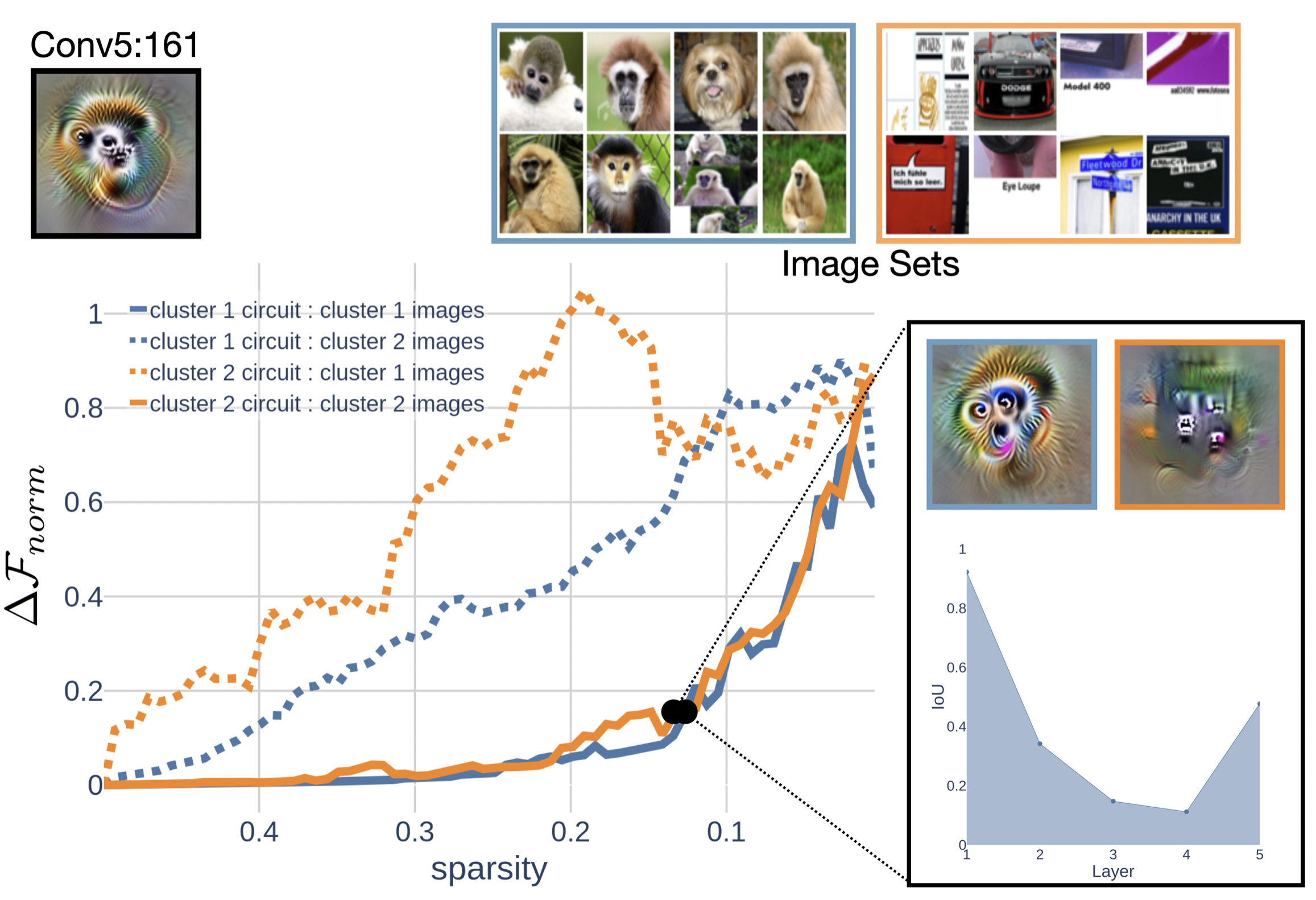}
  \caption{Subcircuit pruning of monkey/text feature.}\label{fig:monkey/text}
\end{figure*}

Even when a circuit is not fully disconnected, it may still contain \textit{dead-end} latent kernels/filters, for which there is no path forward to the target feature or backward to the input. Removing these \textit{dead-end} kernels results in a sparser, yet functionally equivalent circuit. We call the sparsity of a circuit with its dead-ends removed \textit{effective sparsity}. Fig \ref{fig:effective} shows the relationship between circuit sparsity and effective sparsity for all \textit{actgrad} circuits extracted from the regularized model, on a log/log scale. Each point represents a circuit, and the line represents equivalence between the sparsity measures (no dead-ends).
\subsection{Bias only filters}
Finally, when extracting sub-networks for circuit diagrams, one must must reconcile filters' \textit{biases}. Suppose all of a filter's kernels have been pruned from a circuit, we wouldn't want to display such a filter (vertex) in the circuit diagram, but if we simply mask kernels to define the circuit's function, the filter \textit{still} has an effect, as its bias remains. Thus, for circuits displayed as circuit diagrams, if a latent filter has all its kernels masked, we remove the entire filter, bias included. We call circuits with biases \textit{masked} and circuits without biases \textit{pruned}. This bias removal may have an effect on the circuit's feature preservation, thus in Fig \ref{fig:prunevmask} we compare the difference in Pearson R between \textit{masked} and \textit{pruned} circuits (pruned R - masked R), plotted as a function of effective sparsity. As above, we make this comparison for all \textit{actgrad} circuits extracted from the regularized model. Bias removal only has an effect at high sparsities (\(< .1\)), and that effect is usually small. Sometimes bias removal even has a positive effect, leading to a better feature-preserving circuit (dots above the red 0-line).  
\subsection{Hierarchical Sparsity Regularization}
To obtain a regularized version of Alexnet with kernel-wise sparsity, we used hierarchical group sparse regularization \cite{hierarch_reg}, utilizing the associate code, \url{https://github.com/K-Mitsuno/hierarchical-group-sparse-regularization}. We utilized a combination of hierarchical squared group l1/2 regularization \(R_{1/2}\) as well as simple L1 regularization \(R_{1}\). For a given  convolutional layer's weight matrix \(\mathcal{W}_{l}\), these regularization terms are defined as;
\begin{gather} \tag{14}
R_{1 / 2}\left(\mathcal{W}^{l}\right)=\sum_{j=1}^{C_{\text{in}}^{l}}\left(\sum_{i=1}^{C_{\text{out}}^{l}} \sqrt{\sum_{h=1}^{K_{h}^{l}} \sum_{w=1}^{K_{w}^{l}}\left|w_{i, j, h, w}^{l}\right|}\right)^{2}, \\ R_{1}\left(\mathcal{W}^{l}\right)=\sum_{j=1}^{C_{\text{in}}^{l}}\sum_{i=1}^{C_{\text{out}}^{l}} \sum_{h=1}^{K_{h}^{l}} \sum_{w=1}^{K_{w}^{l}}\left|w_{i, j, h, w}^{l}\right|
\end{gather}
These regularization terms are then combined with the typical cross-entropy loss \(\mathcal{L}_{ce}\) to obtain the loss used to train the model;
\begin{equation} \tag{15}
\mathcal{L}=\mathcal{L}_{ce} + \lambda_{2}\sum_{l=1}^{L}\lambda_{1}R_{1 / 2}\left(\mathcal{W}^{l}\right) + (1-\lambda_{2})\sum_{l=1}^{L}\lambda_{1}R_{1}\left(\mathcal{W}^{l}\right) 
\end{equation}
Where \(\lambda_{1}\) scales the regularization terms, and \(\lambda_{2}\) sets the balance between \(R_{1/2}\) and \(R_{1}\). We used \(\lambda_{1}\) = .002 and \(\lambda_{2}\) = .6. We then finetuned a pretrained Alexnet model with this regularizer for 50 epochs using SGD, with a LR=.001, and momentum=.7. The resultant model achieves a top-1 accuracy of 53.03\% on ImageNet, where the original model achieves 56.55\%. From the first convolutional layer to last, the model has the following proportions of kernels \textit{remaining}, as determined by (\(\boldsymbol{S}_{magnitude} > .001\)):
1.0, 0.820, 0.538, 0.634, 0.581.
\subsection{Clustering}
To identify candidate poly-semantic features for our \textit{subcircuit} extraction experiments, we took a data-driven approach. For each filter \(f^{l}_{j}\) in the regularized Alexnet convolutional layers, we identified the highest 300 individual activations it produced in response to the ImageNet training set, under the constraint that no two activations belong to the same activation map (are responses to different parts of the same image). Let \(\mathbf{a}^{l}_{j} = \{a_{j,i}\}_{i=1}^{300}\) refer to these activations, and  \(\mathcal{D}^{l}_{j} = \{\mathbf{x}^{l}_{j,i}\}_{i=1}^{300}\) refer to the corresponding image patches in each activation's effective receptive field. We then consider layer \(l\)'s \textit{population} response to \(\mathcal{D}^{l}_{j}\), a set of 300 \textit{activation vectors} \(\mathcal{P}^{l}_{j}\):
\begin{equation} \label{eq:population vecs}
\mathcal{P}^{l}_{j} := \{p^{l}_{j,i}\}_{i=1}^{300} := 
\{\{f_{h}^{l}(\mathbf{x}^{l}_{j,i})\}_{h=1}^{C_{\text{out}}}\}_{i=1}^{300}
\end{equation}
We reason that for a candidate poly-semantic feature \(f_{j}^{l}\), the set \(\mathcal{P}^{l}_{j}\) should be clusterable. This would suggest \(\mathcal{D}^{l}_{j}\)
is represented as distinct groups in layer \(l\)'s representation space, while the supposed poly-semantic feature \(f_{j}^{l}\) makes no such distinction, as its activations \(\mathbf{a}^{l}_{j}\) are all very high, given our selection criteria for \(\mathbf{a}^{l}_{j}\). 
\par
We used HDBScan clustering \cite{HDBScan} to search for features with clusterable \(\mathcal{P}^{l}_{j}\) (under a BSD 3-Clause license). HDBScan  works well in high-dimensions and requires only a minimum cluster size as its hyperparameter (we used 10). It will return anywhere from 0 to 30 clusters (300/10), allowing us to ignore filters that don't show good clustering, and dissect other filters with respect to potentially many semantic groups (although in practice this approach rarely returned more than 2 clusters). Figure \ref{fig:clusters} shows some examples of feature's with clusterable top-activations identified by this approach.

\subsection{Monkey/text feature}
The splitting of polysemantic features into subcircuits does not only work for \textit{conv5:9}, but rather many such features. We conducted the same analysis outlined in Section 6.1 on another polysemantic feature, \textit{conv5:161}, which is selective for monkey faces and written text (Fig \ref{fig:monkey/text}).
\subsection{Corners with shifts}
\begin{figure}[t]
  \centering
  \includegraphics[width=.8\linewidth]{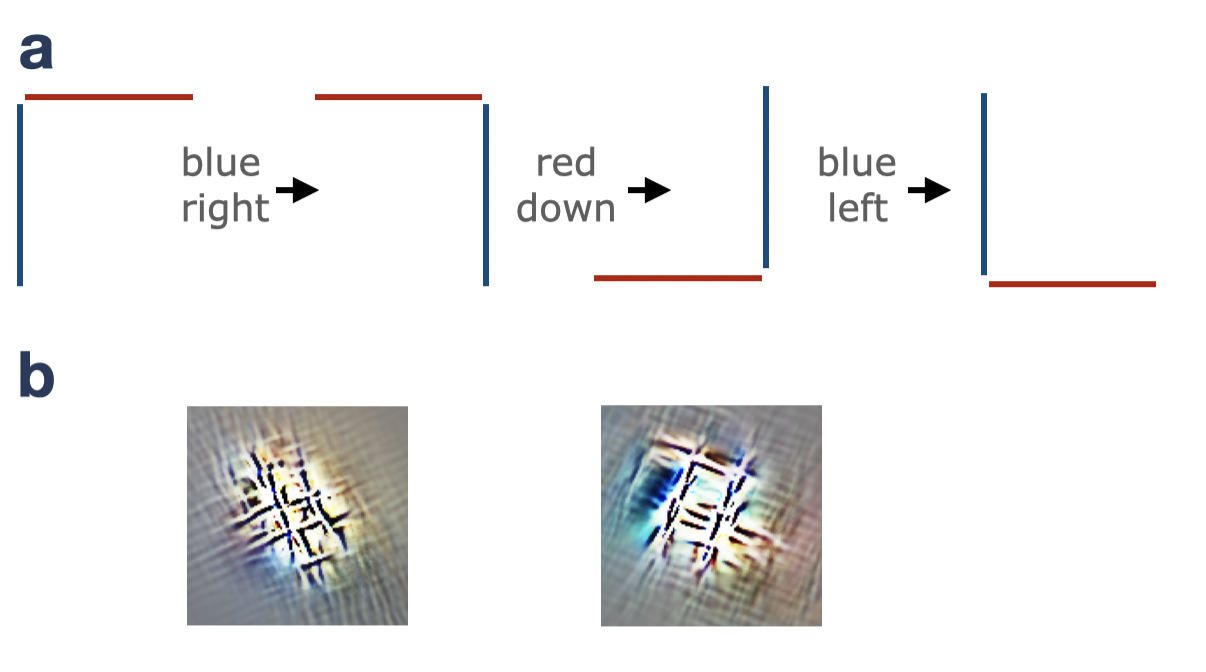}
  \caption{a) Shifting the lines in a corner generates all 90 degree rotations of that corner. b) Negative feature visualizations of two circuits pruned with respect to corners (mod \(45^{\circ}\)).}\label{fig:shift}
\end{figure}

Section 6.2 shows pruning with respect to one corner results in a circuit that preserves its inhibition to all corners mod \(90^{\circ}\). We hypothesize that this can be attributed to the convolution of line detecting kernels. Observe that rotating a corner by \(90^{\circ}\) is equivalent to \textit{translating} one of its edges (Fig \ref{fig:shift}.a). Thus, if circuit pruning based on a corner specifies two such line detectors, the resultant circuit will detect many shifted versions of the two lines. The negative feature visualizations of the 2 corner circuits pruned in section 6.2 show many shifted copies of perpendicular lines. 

\end{document}